\documentclass{article}

\usepackage[preprint]{neurips_2024}

\usepackage[utf8]{inputenc} 
\usepackage[T1]{fontenc}    
\usepackage{hyperref}       
\usepackage{url}            
\usepackage{booktabs}       
\usepackage{amsfonts}       
\usepackage{nicefrac}       
\usepackage{microtype}      
\usepackage{xcolor}         
\usepackage{graphicx}
\usepackage{booktabs}
\usepackage{algorithm}
\usepackage{algpseudocode}

\usepackage[accsupp]{axessibility}  

\usepackage{multirow}
\usepackage{array}
\usepackage{wrapfig}
\usepackage{slashbox}
\usepackage{caption}
\usepackage{subcaption}
\newcolumntype{P}[1]{>{\centering\arraybackslash}p{#1}}

\title{Category Level 6D Object Pose Estimation from a Single RGB Image using Diffusion}

\author{Adam Bethell \\
University of Adelaide \\
\And 
Ravi Garg \\
University of Adelaide \\
\And 
Ian Reid \\
University of Adelaide \\
} 

\begin{document}

\maketitle

\begin{abstract}
  Estimating the 6D pose and 3D size of an object from visual data is a fundamental task in computer vision. Although single-view geometry is a deeply established domain, contemporary category-level methods frequently rely on rigid prerequisites such as precise object models, ground truth depth, or multi-modal LiDAR integration to achieve robust results. In this work, we introduce a unified generative framework that addresses both single-view category-level pose estimation and temporal sequence tracking using only RGB input. Our method leverages score-based diffusion models to generate a rich multi-hypothesis pose distribution, inherently capturing spatial and geometric uncertainties. While existing diffusion-based estimators typically rely on computationally expensive likelihood models to prune outliers, we propose an efficient alternative utilising Mean Shift to directly isolate the distribution's mode as the final pose estimate. Our approach establishes a new state-of-the-art baseline on the challenging REAL275 benchmark among two-stage, crop-based estimators. Furthermore, by decoupling object detection from pose estimation, our generative framework explicitly avoids the catastrophic domain overfitting inherent to end-to-end single-stage detectors, achieving highly robust zero-shot generalisation on the unseen Wild6D dataset. Finally, we demonstrate that the iterative nature of our score-based sampler enables a seamless transition to video sequences by preserving and propagating the multi-hypothesis distribution across time as a coherent temporal prior.
\end{abstract}

\section{Introduction}
\label{sec:intro}
Estimating the 6D object pose and 3D object size from an image or video sequence is a vital task that enables a variety of downstream applications in robotics \cite{du_vision-based_2021, tremblay_deep_2018} and augmented reality \cite{tan_6d_2017, tang_3d_2020}. While early instance-level methods assume a known, CAD model of the target object \cite{shugurov_osop_2022, su_zebrapose_2022, li_deepim_nodate, labbe_cosypose_2020}, category-level pose estimation instead generalises to unseen object instances within a class, and must contend with severe intra-class shape variation. Most category-level methods further assume access to RGB-D sensors or dense point clouds \cite{lin_category-level_2022, zhang_rbp-pose_2022, lin_vi-net_2023, chen_learning_2020}; we address the harder, more broadly deployable setting where pose and size must be recovered from a single RGB image alone, introducing additional scale and depth ambiguity.

Existing RGB-only approaches fall into two paradigms: predicting dense correspondences to a canonical Normalized Object Coordinate Space (NOCS) \cite{wang_normalized_2019} and solving a generalised PnP problem, or directly regressing pose parameters end-to-end. Direct regression is fast but suffers from the multi-hypothesis problem: symmetries, occlusion, and scale ambiguity mean a single image can correspond to multiple valid, disjoint poses, and forcing a network to commit to one output produces unstable training and degraded accuracy even on non-symmetric objects. GenPose \cite{zhang_genpose_nodate} addressed this by learning a score-based diffusion model over point-cloud-conditioned pose hypotheses rather than a single deterministic output, but remains dependent on high-quality 3D input, discards its distributional output during tracking (falling back to single-sample propagation), and relies on a computationally expensive auxiliary network (EnergyNet) to collapse hypotheses into a final estimate.

We introduce a unified generative framework that performs both category-level pose estimation and temporal tracking from RGB alone. Our score-based diffusion model is conditioned on a joint context vector --- an RGB crop, predicted monocular relative depth and surface normals, a category embedding, and a global image feature --- replacing GenPose's 3D point-cloud backbone with dual convolutional encoders operating directly on 2D geometric cues. This sidesteps the scale distortion that back-projected monocular depth otherwise introduces into point-cloud-based pipelines (Supplementary Table 5). We replace EnergyNet's auxiliary filtering-and-mean-pooling with Mean Shift \cite{comaniciu_mean_2002} mode-seeking, which converges directly onto the empirical distribution's density peak rather than averaging across disjoint symmetry modes --- eliminating a second network while improving both accuracy and speed. Finally, because our framework retains a full hypothesis set rather than a single point estimate at every step, it extends naturally to tracking: we propagate the entire multi-hypothesis distribution across frames as a structured temporal prior, rather than collapsing to one pose before re-perturbing, which helps resist mode collapse under occlusion or rapid motion.

We evaluate our framework on the REAL275 benchmark and perform zero-shot inference on the Wild6D dataset to rigorously test cross-domain generalisation. We specifically scope our framework within the class of two-stage, crop-based RGB estimators, which explicitly decouple 2D detection from 3D pose extraction to prevent catastrophic domain overfitting. Among these architectures, our approach establishes a new state-of-the-art on the strict joint \(10^{\circ}10cm\) metric. By trading the in-domain absolute scale bias inherent to uncropped single-stage models for calibrated multi-hypothesis uncertainty, our generative formulation unlocks seamless temporal tracking and highly resilient zero-shot generalisation.

Our contributions are as follows:

\begin{itemize}
    \item \textbf{A generative RGB-only framework for joint pose estimation and tracking:} We adapt score-based diffusion to a 2D semantic-geometric context vector, resolving scale ambiguity without depth sensors, CAD models, or shape priors.
    \item \textbf{Distribution-aware temporal tracking:} We propagate the full multi-hypothesis pose distribution across frames as a temporal prior, rather than a single collapsed estimate, improving robustness to occlusion and symmetric ambiguity.
    \item \textbf{Efficient mode-seeking via Mean Shift:} We reformulate hypothesis aggregation as maximum-likelihood mode-seeking, removing the need for an auxiliary energy network while improving both accuracy and inference speed.
    \item \textbf{State-of-the-art performance on REAL275:} We demonstrate state-of-the-art results among two-stage, generative RGB-only methods on REAL275, alongside strong zero-shot generalisation to Wild6D.
\end{itemize}

\section{Related Work}
\subsection{RGB-D Based Category Level Object Pose Estimation}
The task of category-level object pose estimation is to predict the object poses of novel objects from a certain classes. Normalised Object Coordinate Space (NOCS) \cite{wang_normalized_2019} approaches the task by introducing a normalised canonical coordinate space for each category and uses this space to generate 3D-3D correspondences from which the Umeyama algorithm \cite{umeyama_least-squares_1991} can estimate the pose. To estimate the NOCS representation of the object the notion of Shape Prior Deformation is introduced by \cite{tian_shape_2020} where a category shape prior is used as input to output a deformation field.

SGPA \cite{chen_sgpa_2021} extends this idea to transformers, leveraging the attention mechanism between the image, depth map and shape prior. The NOCS representation is also used as additional supervision by \cite{wang_category-level_2021}, \cite{lin_category-level_2022} and \cite{zhang_ssp-pose_2022} while directly regressing the pose from the RGB-D image. As previously mentioned the regression methods inherently suffer from the multi-hypothesis issue so \cite{lin_sar-net_2022} and \cite{di_gpv-pose_2022} design symmetry-aware loss functions to address the issue. Similar to the iterative deformation method \cite{wang_category-level_2021}, \cite{chen_category_2020} use an analysis by synthesis style pose estimation method where a VAE and warping operations are used to generate images from a given pose and a segmented image. 

\subsection{RGB Based Category Level Object Pose Estimation}
The analysis by synthesis method presented in \cite{chen_category_2020} is also trained for RGB images and introduces a baseline for RGB-based category-level object pose estimation. While this method produced accurate rotations, there were high translation errors due to scale ambiguity. CPS++ \cite{manhardt_cps_nodate} trains a regression network for object pose and size estimation using synthetic data with an additional refinement stage using depth for real domain transfer. The scale ambiguity present in the analysis by synthesis method is addressed in MSOS \cite{lee_category-level_2021}, OLD-Net \cite{fan_object_2022} and DMSR \cite{wei_rgb-based_2023} using different networks or branches for the NOCS representation and the object scale. LaPose \cite{leonardis_lapose_2025} follows these and predicts the correspondences using a Laplacian distribution. The most recent work GIVEPose \cite{huang_givepose_2025} proposes a new coordinate correspondence map for category-level consensus. 

Further advancing RGB-only architectures, RCGNet \cite{rcgnet} introduces a transformer-based network equipped with a geometric feature-guided algorithm to extract object geometry without the need to explicitly predict depth maps. To tackle the limitations of multi-stage pipelines, YOPO \cite{lee2026yopo} proposes a minimalist single-stage detection transformer that directly performs 2D detection and 9D pose regression. Similarly, Fischer et al. \cite{fischer2025unifiedcategorylevelobjectdetection} present a single-stage approach using 3D prototypes and neural mesh models combined with multi-model RANSAC. Additionally, LanCOPE \cite{LanCope} integrates language descriptions through a cross-modal differential perception feature fusion network to better guide the extraction of robust category features. In contrast to these deterministic regression or dense correspondence methods, we do not require correspondence or symmetry supervision or separate scale networks, instead relying on the distribution modelling capabilities of diffusion models.

\subsection{Diffusion Models}
Score-based generative models have advanced data distribution modelling by estimating the log-likelihood gradient via Denoising Score Matching (DSM) \cite{hyvarinen_estimation_nodate, song_generative_2019, vincent_connection_2011}. Their flexibility with conditional inputs has enabled frameworks like GenPose \cite{zhang_genpose_nodate}, which generates pose hypotheses from point clouds, relying on energy-based filtering and mean pooling for the final estimate.

Recently, diffusion paradigms have been adapted for category-level pose estimation. MonoDiff9D \cite{MonoDiff9D} utilises DINOv2 to estimate a coarse point cloud from an RGB image in a zero-shot manner, mapping this geometric information into conditional encoding for a transformer-based denoiser. Addressing domain gaps, Diff9D \cite{Diff9D} frames domain-generalised pose estimation as a generative process trained purely on synthetic data; however, it strictly requires depth information (point clouds) obtained from an RGB-D camera.

In this work, we take inspiration from this generative approach but introduce three critical modifications to improve robustness and efficiency. First, we adapt the score-based architecture to operate on RGB images rather than point clouds. Second, we move away from energy-based filtering and mean pooling, opting instead to employ Mean Shift to identify the mode of the distribution as our final pose estimate, which provides a more robust aggregation strategy. Finally, we extend the GenPose \cite{zhang_genpose_nodate} tracking paradigm by explicitly propagating the full multi-hypothesis distribution across frames, utilising it as a structurally rich temporal prior that informs subsequent tracking.

\begin{figure*}[t]
  \centering
    \includegraphics[height=4cm]{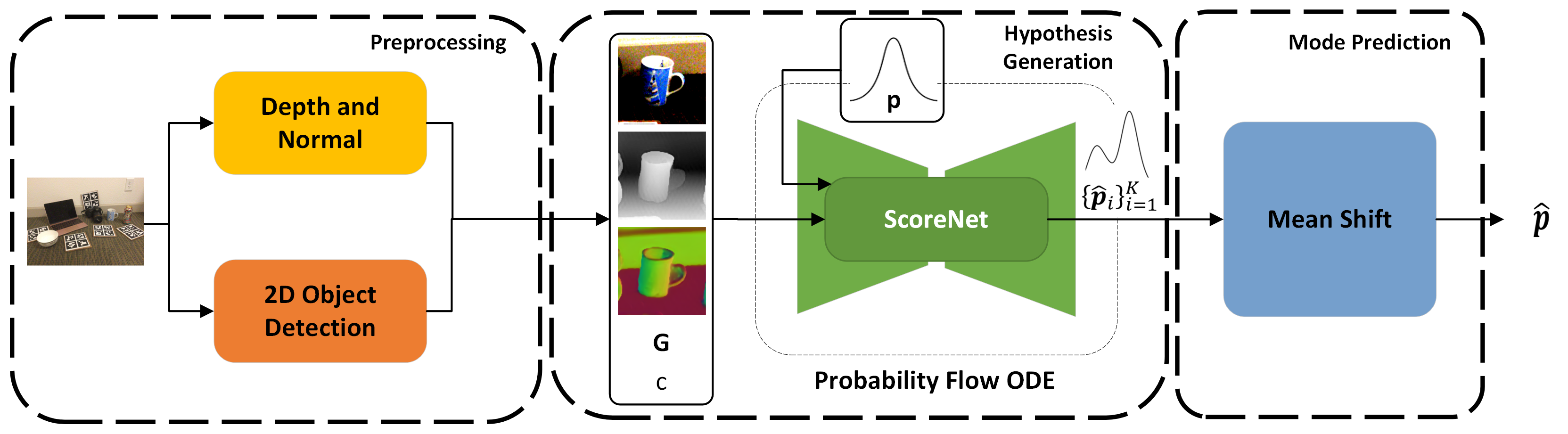}  
    \caption{An overview of our method. \textbf{Preprocessing:} A depth and normal predictor and 2D object detector used to get the cropped images, global image feature \(\textbf{G}\) and category ID \(c\). \textbf{Hypothesis Generation:} Pose hypotheses refined using a ScoreNet to generate the final pose hypotheses \(\{\hat{\mathbf{p}}\}_{i=0}^K\). \textbf{Mode Prediction:} Mean Shift on the pose hypotheses for the final pose estimate \(\hat{\mathbf{p}}\).}
  \label{fig:overview}
\end{figure*}

\section{Method}
In this work, our goal is to estimate object pose and scale, denoted by \(\mathbf{p} = \{ \mathbf{R} \in SO(3), \mathbf{t} \in \mathbb{R}^3, \mathbf{s} \in \mathbb{R}^3 \}\). Our system is based on learning a score function to denoise pose hypotheses via diffusion. Training this function utilizes a dataset \(\mathcal{D}\) comprising RGB images, object categories, and ground-truth poses: \(\mathcal{D} = \{ (I_i, c_i, \mathbf{p}_i) \}_{i=1}^n\), where \(I_i \in \mathbb{R}^{H \times W \times 3}\) represents the input images, \(c_i \in \{ 1, \dots, L \}\) denotes the category ID, and \(\mathbf{p}_i\) is the ground-truth pose. At inference time, our objective is to recover the pose \(\mathbf{p}\) given a single RGB image \(I\).

As illustrated in Figure \ref{fig:overview}, our pose estimation pipeline comprises three stages: feature-extracting backbone, pose hypothesis generation, and mode prediction. We detail each of these steps below, highlighting the justifications for our critical design choices.

\subsection{Network Architecture and Pre-processing}
\label{subsec:architecture}
We extract object bounding boxes, category ID \( c \) and a global image feature \(G\) from a 2D object detector, namely MaskRCNN trained on COCO \cite{lin_microsoft_2015}, which was finetuned on the target pose estimation dataset by \cite{wang_normalized_2019}. Additionally, as in \cite{wei_rgb-based_2023}, we use a pretrained model from Omnidata \cite{eftekhar_omnidata_2021, kar_3d_nodate} to extract the relative depths and normals of the object. While \cite{wei_rgb-based_2023} opts to estimate the depth and normal from the cropped image of the object using the estimated bounding box, we estimate them from the full image and use Dynamic Zoom In (detailed in the supplementary material) to crop them. The dataset \(D\) is thus augmented with a cropped image \(I\), cropped relative depth \(I_D\), cropped normals \( I_n\), category ID \(c\), and global feature \(G\).

The architecture of our ScoreNet comprises two CNN backbones encoding the cropped RGB image of the object \(I\) and a four-channel image consisting of concatenated relative depths \(I_D\) and normals \(I_n\) respectively as in DMSR \cite{wei_rgb-based_2023}. We use sinusoidal positional encoding of timestep \(t\) as the time embedding and learned encodings for the Category IDs. The output embeddings mentioned above are concatenated with the global image feature \(G\) to form the conditioning vector, $\mathcal{C}$ for diffusion. We encode the 12D pose vector via an MLP to a 256-dimensional vector for denoising and use four separate decoder heads to estimate rotation (two heads for continuous 6D representation), scale-invariant translation (SITE) \cite{li_cdpn_2019}, and 3D size respectively. Finally, we introduce a scale-invariant representation for the 3D size head, ensuring that our model's object dimension predictions remain decoupled from and invariant to the geometric scaling artifacts introduced by the 2D bounding box cropping step.

\subsection{Multi-Hypothesis Diffusion Generation}
\label{subsec:diffusion}
It has been well established that pose estimation methods that heavily rely on regression suffer from the multi-hypothesis issue due to object symmetries and occlusions. To address this, we choose to use multiple-pose hypothesis generation by adapting the score-based diffusion modelling strategy of GenPose \cite{zhang_genpose_nodate}. Crucially, we remove the conditioning on the 3D point cloud used in GenPose and instead condition our ScoreNet, $\mathbf{s}_\mathbf{\theta}$ on a joint context vector $\mathcal{C} = \{I, I_D, I_n, c, G\}$, which aggregates the RGB image, relative depths, surface normals, object category, and global image features.

This results in a streamlined loss function for training our diffusion model \(\mathbf{s}_\mathbf{\theta}\):
\begin{equation}
 L(\theta) = \mathbb{E}_{t, \mathbf{p}(0), \mathbf{p}(t)} \left[ \lambda(t) \left\| \mathbf{s}_\mathbf{\theta}(\mathbf{p}(t), \mathcal{C}, t) + \frac{\mathbf{p}(t) - \mathbf{p}(0)}{\sigma(t)^2} \right\|^2_2 \right]
  \label{eq:loss}
\end{equation}
where \(t \in [0,1]\) is the continuous time variable used to generate the perturbed pose \(\mathbf{p}(t)\) sampled using the Variance-Exploding Stochastic Differential Equation (VE-SDE) introduced in \cite{song_score-based_2021}. We specifically adopt the continuous-time VE-SDE formulation rather than discrete-time DDPMs because it allows for highly flexible step-size scheduling during the Probability Flow ODE reverse sampling, which is critical for aggressively scaling down denoising steps to meet the near-real-time requirements of continuous pose tracking.

Once the ScoreNet has been trained, we utilise the standard generative inference procedure from GenPose \cite{zhang_genpose_nodate} to sample the discrete pose hypotheses. Specifically, a deterministic Probability Flow ODE is used to reverse-denoise a set of random initial vectors into the final hypothesis set \(\{\hat{\mathbf{p}}\}_{i=1}^K\). We emphasise that while our conditioning architecture and network inputs are novel (Sec.~\ref{subsec:architecture}), the underlying ODE sampling mechanism remains unmodified; our proposed pipeline modifications focus strictly on how these generated hypotheses are aggregated (Sec.~\ref{subsec:mode_seeking}) and temporally propagated (Sec.~\ref{sec:pose_tracking}).

One can argue that GenPose \cite{zhang_genpose_nodate} can be trivially adapted for RGB-only pose estimation by replacing its ground-truth point cloud input with a 3D point cloud back-projected from a single-view metric depth estimator. We show, however, that such an adaptation grossly underperforms our approach, as presented in Table 5 of the Supplementary. This performance gap stems from a fundamental architectural constraint: GenPose relies on a 3D coordinate-based backbone (PointNet++) to process its inputs. Because single-view metric depth estimators—such as UniDepth \cite{piccinelli_unidepth_2024} used in this baseline experiment—are prone to significant global scale errors, the resulting back-projected point clouds suffer from spatial warping. This spatial distortion misaligns the coordinates used in PointNet++, causing the geometric encoding to degrade. 

This architectural limitation motivates our shift to processing 2D relative depth maps ($I_D$) and surface normals ($I_n$) via fully convolutional image backbones. By decoupling feature extraction from absolute 3D coordinate spaces, our network bypasses early scale sensitivity entirely. The correct metric scale is instead resolved downstream by the diffusion model combining the global image feature $G$, the category ID $c$ and the image features. 
Interestingly, DMSR \cite{wei_rgb-based_2023} uses an explicit object shape prior to assist in pose estimation and help guide scale recovery. However, we argue that learning complex, category-specific priors for full 3D object geometry is overkill; a direct, learned mapping from category ID and global features directly to pose dimensions suffices. While our category ID embeddings function as a strong, coarse prior for absolute scale, the global image feature $G$ provides critical spatial context that allows the model to recover information lost due to object cropping and scaling operations. Crucially, because these conditioning features act as structural priors directly on the output regression heads rather than trying to un-warp an intermediate 3D point cloud structure, they function as a highly robust mechanism for scale recovery, with these inputs being dropped during training following the classifier-free guidance detailed in the supplementary material.

\subsection{Mode Seeking via Mean Shift}
\label{subsec:mode_seeking}
While the ScoreNet provides a robust hypothesis set \(\{\hat{\mathbf{p}}\}_{i=1}^K\), downstream tasks typically require a single pose prediction. In GenPose \cite{zhang_genpose_nodate}, the final pose estimate is acquired via a two-stage filtering and aggregation process. First, an auxiliary energy model (EnergyNet) is trained using a surrogate objective to predict the unnormalised log-likelihood (energy score) of each hypothesis given a 3D point cloud. Hypotheses are sorted using their energies to filter out the last \((1 - \delta \%)\) with the remaining samples aggregated using unweighted mean pooling operation. 

A possible approach for GenPose would be to compute a weighted average of the hypotheses directly, using the predicted EnergyNet scores as probability weights. However, in practice, the raw energy scores produced by such networks are highly uncalibrated and non-linear, meaning a soft weighted average is easily biased by overconfident outlier hypotheses. Consequently, GenPose resorts to a filtering followed by an unweighted mean. This approach remains fundamentally flawed, as mean pooling over the surviving samples collapses to an incorrect expected value when dealing with multi-modal distributions. In particular, this creates a severe mean-mode discrepancy when an object possesses discrete symmetrical properties, as averaging across separate geometric modes yields an unfeasible pose.

To circumvent both the computational burden of training an auxiliary EnergyNet and the theoretical pitfalls of mean pooling, we propose a clean maximum likelihood formulation. Instead of attempting to weight hypotheses using poorly calibrated raw network scores, we treat the sampled set \(\{\hat{\mathbf{p}}\}_{i=1}^K\) as an empirical distribution. We utilize the mode-seeking capabilities of Mean Shift \cite{comaniciu_mean_2002, yizong_cheng_mean_1995} to isolate the analytical peak of the pose distribution. Crucially, because Mean Shift seeks localised density modes based on geometric clustering rather than individual sample weight metrics, it is inherently robust to outliers and eliminates the need for any secondary likelihood-prediction network. For our implementation, Mean Shift++ \cite{jang_meanshift_2021} is executed on the rotation, translation, and size components independently, selecting the mode associated with the largest localised cluster to construct our final pose prediction.

\begin{table*}[t]
  \centering
  \resizebox{\textwidth}{!}{%
  \begin{tabular}{l c c c c c c c c c c c}
    \toprule
    Method & AUC IoU & Mean IoU & AUC Rot & Mean Rot & AUC Trans & Mean Trans & IoU25 & IoU50 & $10^\circ$ & 10cm & $10^\circ$10cm \\
    \midrule
    DMSR \cite{wei_rgb-based_2023}      & 22.1 & 45.3 & 18.3 & 11.6 & 40.2 & 6.4 & 37.4 & 16.3 & 36.9 & 40.0 & 25.2 \\
    MonoDiff9D \cite{MonoDiff9D}        & 19.7 & 39.6 & 64.0 & 10.6 & 46.6 & 11.9 & 36.3 & 11.1 & 39.3 & 45.5 & 27.7 \\
    LaPose \cite{leonardis_lapose_2025} & 23.7 & 43.2 & 58.9 & 7.44 & 45.5 & 6.46 & 41.2 & 17.5 & 47.5 & 44.4 & 30.5 \\
    GIVEPose \cite{huang_givepose_2025} & 25.1 & 44.9 & 62.7 & 6.24 & 47.1 & 6.41 & 42.9 & 20.1 & 53.1 & 45.9 & 34.2 \\
    \midrule
    Ours & \textbf{25.7} & \textbf{46.4} & \textbf{65.2} & \textbf{5.87} & \textbf{50.7} & \textbf{5.7} & \textbf{47.0} & \textbf{22.9} & \textbf{61.3} & \textbf{54.6} & \textbf{47.0} \\
    \bottomrule
  \end{tabular}%
  }
  \caption{Quantitative comparison of cropped RGB-based category-level object pose estimation on the REAL275 dataset.}
  \label{tab:sota}
\end{table*}

\subsection{Pose Tracking Algorithm}
\label{sec:pose_tracking}
While static single-frame estimation provides a strong baseline, video sequences offer temporal context that can be leveraged to resolve ambiguities arising from occlusion or rapid camera movement. Unlike conventional methods that collapse the distribution of the previous frame $t-1$ into a single pose vector $\hat{\mathbf{p}}_{t-1}$ before adding Gaussian perturbation for frame $t$, our framework explicitly propagates the entire multi-modal distribution over time. This design prevents early mode collapse, which is vital when navigating the complex, multi-hypothesis landscape inherent to category-level RGB input.

Let $P(\mathbf{p}_{t-1} | I_{1:t-1})$ denote the continuous multi-modal distribution over the 9D object pose and 3D scale at frame $t-1$, represented by our set of $K$ generated hypotheses $\mathcal{H}_{t-1} = \{\hat{\mathbf{p}}_{t-1}^i\}_{i=1}^K$. Instead of initialising the score-based diffusion model at frame $t$ with completely random noise or a single perturbed state, we define a distribution-aware transition prior. 

The tracking process proceeds via the following stages:
\begin{enumerate}
    \item \textbf{Temporal Propagation:} The hypothesis set $\mathcal{H}_{t-1}$ from the previous frame is propagated forward using a constant-velocity or random-walk motion model to form a prediction set $\hat{\mathcal{H}}_t$. Rather than destroying the underlying distribution structure, this step preserves the geometric cluster topography corresponding to distinct pose modes or object symmetries.
    
    \item \textbf{Diffusion Conditioning:} The score-based diffusion process for the current frame $t$ is conditioned on $\mathcal{C}$, alongside the propagated distribution $\hat{\mathcal{H}}_t$ using a smaller noise level than single-frame prediction. This transition prior guides the reverse denoising steps, acting similarly to a particle filter motion model but refined within our ScoreNet architecture.
    
    \item \textbf{Multi-Hypothesis Refinement:} The reverse diffusion steps gradually drive the random components of the prediction toward the high-density regions of the posterior distribution $P(\mathbf{p}_t | I_{1:t})$. Because our framework maintains the full distribution across frames, it retains the capacity to simultaneously track separate, equally plausible pose hypotheses (e.g., a $180^\circ$ symmetry ambiguity) until subsequent frames provide enough visual evidence to resolve the ambiguity.
\end{enumerate}

By avoiding a premature hard decision on the true pose during frame-to-frame transitions, our tracking mechanism ensures robust temporal consistency and prevents catastrophic tracking failure in challenging, unconstrained video streams. Further details of the tracking algorithm are presented in the supplementary material.

\begin{figure*}[t]
  \centering
  \includegraphics[height=3.5cm]{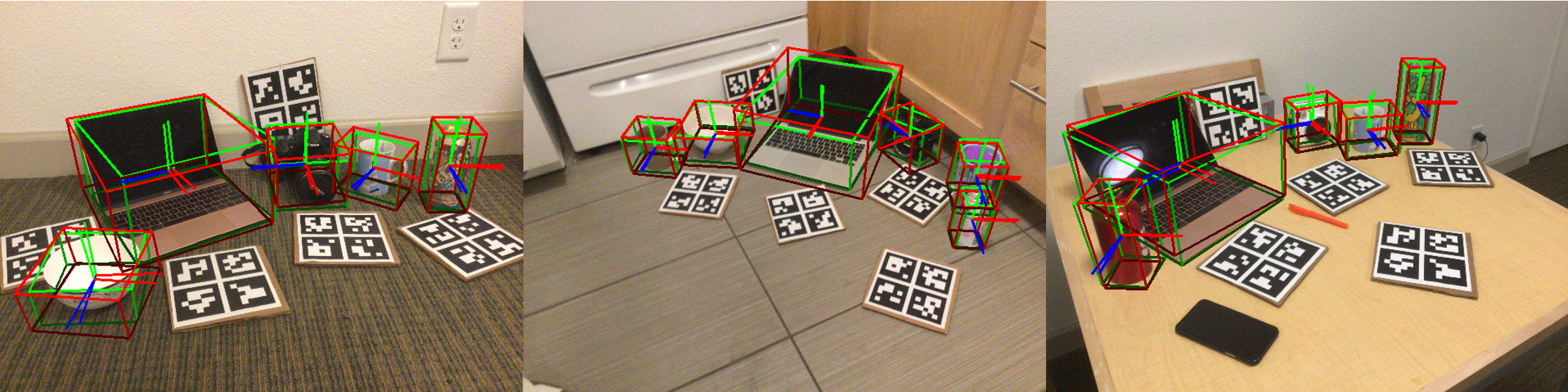}  
  \caption{Qualitative results of our method on the REAL275 dataset.}
  \label{fig:qual}
\end{figure*}

\subsection{Implementation Details}
\label{subsec:implementation}
We generate initial 2D object crops using MaskRCNN \cite{he_mask_2017} and extract relative depth and surface normals via Omnidata \cite{eftekhar_omnidata_2021, kar_3d_nodate}. Our ScoreNet processes these inputs utilising ResNet-34 \cite{he_deep_2015} backbones; comprehensive architectural and training details are provided in the Supplementary Material.

\section{Experiments}
\subsection{Datasets}
We conduct our experiments on two benchmark datasets for 6D category-level object pose estimation introduced by NOCS \cite{wang_normalized_2019}: CAMERA and REAL275. The datasets contain 6 categories of common objects: bottle, bowl, camera, can, laptop and mug. The CAMERA dataset contains synthetic images generated using CAD models of the objects on real background images with 275K training images. The REAL275 dataset contains 3 unique instances of each category in the real world with 7 different scenes each containing a minimum of 5 objects per scene. In total, the REAL275 dataset contains 4300 training images and 2750 testing images. 

\subsection{Metrics}
Following NOCS \cite{wang_normalized_2019}, we evaluate pose and size estimation using Mean Average Precision (mAP) and 3D Intersection-Over-Union (IoU) as well as Area Under the Curve (AUC), and mean errors. We report mAP at strict rotation and translation thresholds ($10^\circ$, 10cm, and joint $10^\circ10\text{cm}$) and 3D IoU for bounding box overlaps at 25\% and 50\%. To comprehensively capture geometric alignment across the predictive distribution, we also provide the AUC and mean error for each pose component. For symmetric objects (bottle, bowl, can, and handle-occluded mugs), rotation errors around the axis of symmetry are explicitly ignored \cite{wang_normalized_2019}.

\begin{table}[tb]
  \centering
  \begin{tabular}{|P{6em}| P{2em} P{2em} P{2em} P{2em}  P{4em}|}
    \toprule
    Method & IoU25 & IoU50 & \(10^o\) & \(10cm\) & \(10^{\circ}10cm\) \\
    \midrule
    LaPose &  2.3 & 0.1 & 3.1 & 1.1 & 0.1 \\
    GIVEPose &  18.5 & 9.2 & 8.5 & 9.4 & 1.5 \\
    MonoDiff9D &  19.1 & 7.4 & 14.9 & 14.0 & 4.6 \\
    Ours & \textbf{20.1} & \textbf{11.3} & \textbf{24.7} & \textbf{36.7} & \textbf{18.3} \\
  \bottomrule
  \end{tabular}
  \caption{Zero Shot Results on the WILD6D dataset.}
  \label{tab:wild}
\end{table}

\subsection{Comparison with State-of-the-Art}
Table \ref{tab:sota} presents the quantitative evaluation of our framework against state-of-the-art (SOTA) cropped RGB-based estimators on the REAL275 dataset. Our method establishes a new state-of-the-art across all evaluated geometric and spatial metrics.
\begin{itemize}
\item Holistic Geometric Alignment (AUC \& Means): Capturing the full predictive distribution yields superior continuous accuracy. Our framework achieves the highest AUC across all dimensions, reaching 65.2 for rotation and 50.7 for translation. It simultaneously drives down overall prediction drift, setting new benchmark lows for mean rotation error ($5.87^\circ$) and mean translation error (5.7cm).
\item Strict Pose Thresholds: We achieve a massive 37\% improvement on the $10^\circ10\text{cm}$ metric, jumping from GIVEPose's 34.2 to 47.0. Individual threshold accuracies also see significant gains, with rotation ($10^\circ$) increasing by 15\% and translation (10cm) by 19\% over previous baselines.
\item Spatial Overlap (3D IoU): Our decoupled generative formulation improves overall spatial bounding box estimation. We surpass GIVEPose by over 9\% on $\text{IoU}_{25}$ (achieving 47.0) and set a new peak $\text{IoU}_{50}$ of 22.9, while pushing the continuous Mean IoU to a field-leading 46.4.
\end{itemize}
 
Figure \ref{fig:qual} shows the estimated and ground truth poses for images from the REAL275 dataset. The estimated poses for the symmetric objects generally only differ from the ground truth along the symmetric axis further demonstrating the capability of our method to capture the multiple feasible poses. More qualitative results are presented in the supplementary. 

\begin{figure}[tb]
  \centering 
  \begin{subfigure}[b]{0.24\linewidth}
    \centering
    \includegraphics[width=\linewidth]{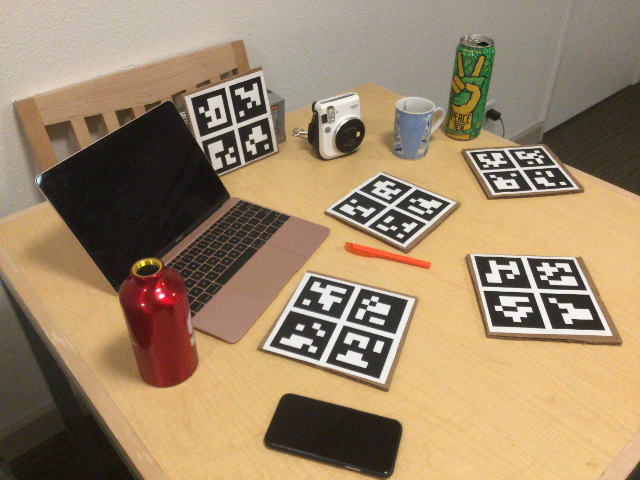}
    \caption{Input image}
    \label{fig:dist-a}
  \end{subfigure}\hfill
  \begin{subfigure}[b]{0.24\linewidth}
    \centering
    \includegraphics[width=\linewidth]{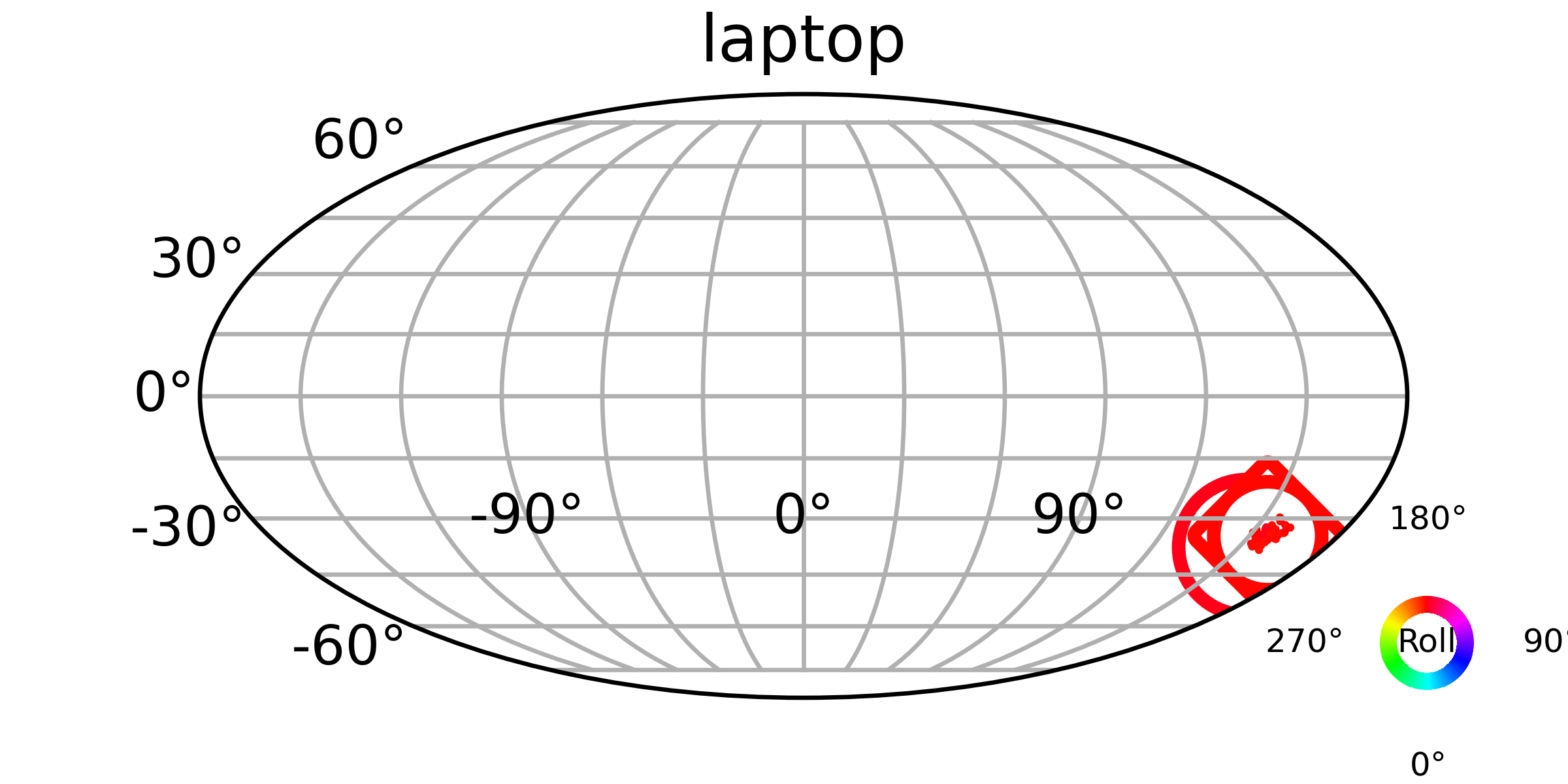}
    \caption{Laptop rotations}
    \label{fig:dist-b}
  \end{subfigure}\hfill
  \begin{subfigure}[b]{0.24\linewidth}
    \centering
    \includegraphics[width=\linewidth]{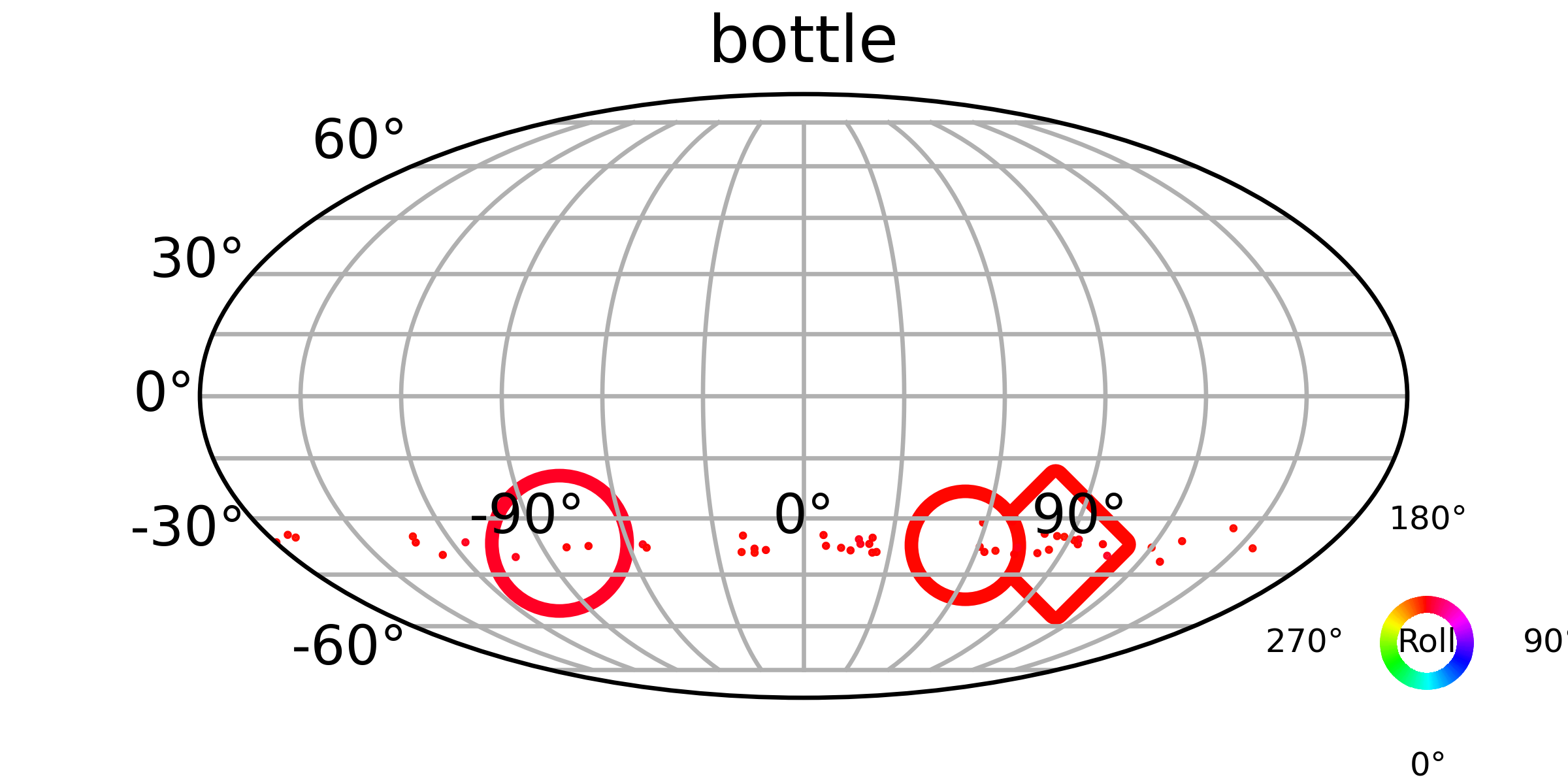}
    \caption{Bowl rotations}
    \label{fig:dist-c}
  \end{subfigure}\hfill
  \begin{subfigure}[b]{0.24\linewidth}
    \centering
    \includegraphics[width=\linewidth]{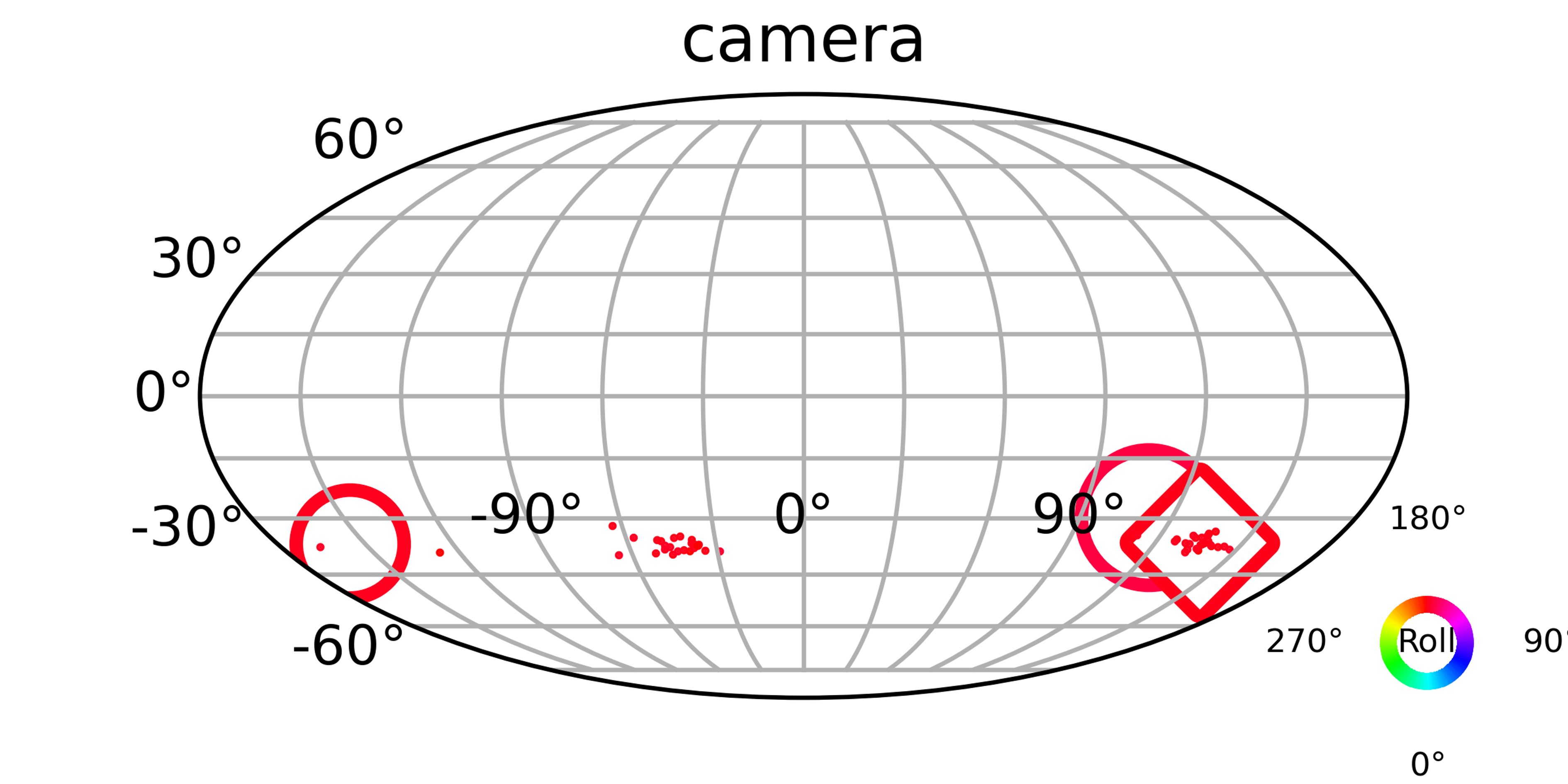}
    \caption{Camera rotations}
    \label{fig:dist-d}
  \end{subfigure}
  \caption{Visualisation of the predicted rotation distribution. Yaw and pitch are mapped to latitude and longitude, respectively, while roll is represented by colour. The ground truth pose is indicated by the large circle, and our final estimate (the mode) is marked with a diamond. Individual pose hypotheses are shown as dots, and the mean pose is depicted as a small circle.}
  \label{fig:dist}
\end{figure}

Furthermore, the qualitative distribution landscapes in Figure~\ref{fig:dist} validate our framework's capacity to capture complex probability distributions. For asymmetric objects, the model produces tight, highly localised hypothesis clusters as seen in Figure~\ref{fig:dist-b}. Conversely, for symmetric objects, our model preserves the true underlying geometry by allowing hypotheses to spread continuously along the axis of symmetry, as clearly depicted in Figure~\ref{fig:dist-c}. 

Most notably, the hypothesis distribution for the \textit{camera} category, shown in Figure~\ref{fig:dist-d}, underscores the primary advantage of our mode-seeking approach. Here, the pose distribution is multi-modal, reflecting the model's uncertainty regarding the camera’s forward versus backward orientation. While a simple mean of the distribution results in an implausible pose that lies outside the manifold of valid orientations—effectively 'averaging out' the modes—our framework successfully captures the underlying density to pinpoint the optimal pose without requiring an auxiliary energy model. As illustrated by the mean-pooling baseline (represented by the smaller circle falling entirely outside the high-density regions predicted by the ScoreNet), standard expectation-based aggregation collapses in multi-modal spaces. By converging directly onto the true analytical peak of the empirical distribution instead, our Mean Shift pipeline completely bypasses this mean-mode discrepancy, resulting in a more accurate and geometrically valid pose estimate.

\subsubsection{Generalisation to Unseen Datasets}
To evaluate the generalisability of our framework to entirely unseen environments, we perform zero-shot inference on the Wild6D dataset. As demonstrated in Table \ref{tab:wild}, our method exhibits superior robustness across both spatial intersection and pose alignment metrics without any fine-tuning on the target domain. For 3D bounding box overlap, we achieve 20.1 and 11.3 on IoU$_{25}$ and IoU$_{50}$ respectively, consistently outperforming the nearest baselines MonoDiff9D (19.1 and 7.4) and GIVEPose (18.5 and 9.2).

This generalisation gap widens significantly when evaluating strict geometric alignment. Our framework reaches 24.7 for rotation accuracy ($10^\circ$) and 36.7 for translation accuracy (10cm), more than doubling the translation performance of MonoDiff9D (14.0). Culminating in 18.3\% on the strict joint \(10^{\circ}10cm\) metric, our approach dramatically eclipses MonoDiff9D (4.6) and GIVEPose (1.5\%). This sweeping performance confirms that our distribution-aware generative process successfully learns robust, transferable geometric representations, preventing the catastrophic overfitting to NOCS camera intrinsics or specific background distributions seen in prior methods.

\

\begin{table}[h]
  \centering
  \begin{tabular}{|P{4em} | P{3em} P{3em} P{4em}|P{3em}|}
    \toprule
    Method & \(10^o\) & \(10cm\) & \(10^{\circ}10cm\) & FPS\\
    \midrule
    LaPose  & 80.1 & 57.1 & 44.7& 9.1\\
    GIVEPose  & 84.9 & 61.9 & 51.6 & \textbf{10.4}\\
    \midrule
    Ours (S) & 86.0 & 62.9 & 59.7 & 6.8  \\
    Ours (M)  & 86.8 & 64.2 & 60.5 & 6.8\\
    Ours (F) & \textbf{87.7} & \textbf{67.5} & \textbf{61.1} & 3.6\\
  \bottomrule
  \end{tabular}

  \caption{Accuracy results of category-level object pose tracking on the REAL275 dataset. We compare our method variants: (S) using a single previous pose for initialisation, (M) leveraging the full pose distribution, and (F) performing restarts for every frame.}
  \label{tab:track}
\end{table}

\begin{table*}[tb]
  \centering
  \begin{tabular}{|P{6em} P{6em}| P{2em} P{2em} P{2em} P{2em} P{5em}| P{3em} P{3em}|}
    \toprule
    Filtering & Aggregation & IoU25 & IoU50 & \(10^o\) & \(10cm\) & \(10^{\circ}10cm\) & S-FPS & T FPS \\
    \midrule
    None  &  Mean Pooling & 46.8 & 20.2 & 60.5 & 54.2 & 46.0 & \textbf{3.6} & 6.7 \\
    Energy  & Mean Pooling & 46.9 & 21.2 & 60.8 & 54.5 & 46.4 & 3.1 & 3.6  \\
    Energy & Mean Shift & \textbf{47.2} & \textbf{23.4}   & \textbf{61.3} &\textbf{55.1} & \textbf{47.2} & 3.1 & 3.7\\
    None & Mean Shift & 47.0 & 22.9   & \textbf{61.3} & 54.6 & 47.0 & \textbf{3.6} & \textbf{6.8}\\
  \bottomrule
  \end{tabular}
  \caption{Ablation on the effectiveness of Mean Shift on the REAL275 dataset with the Frames per second for static pose estimation (S-FPS) and pose tracking (T-FPS).}
  \label{tab:rank}
\end{table*}

\subsection{Category-level Object Pose Tracking}
Before detailing the tracking performance, it is critical to clarify the evaluation protocol, as the 2D detection pipeline utilised in this section differs from the static single-frame evaluation presented in Section 4.3. For the static evaluation (Table 1), our pipeline processes all raw 2D bounding box detections, and the final pose predictions are selected post-hoc based on their 3D IoU overlap with the ground truth bounding boxes. While standard, this exposes the pose estimation metrics to the inherent noise, jitter, and false positives of the raw 2D detector. Conversely, the primary objective of the tracking evaluation (Table 2) is to assess the temporal multi-hypothesis tracking mechanisms in isolation. To facilitate stable frame-to-frame association and remove the necessity of complex 2D-to-3D matching during evaluation, the 2D bounding boxes in the tracking setting are pre-aligned and filtered prior to pose estimation, ensuring only the most accurate 2D bounding box is provided for each object. This decoupled evaluation isolates the geometric precision of the pose estimators from 2D detector instability, explicitly accounting for the performance discrepancy between our static method (47.0 on \(10^{\circ}10cm\)) and the non-tracking fallback Ours (F) (61.1 on \(10^{\circ}10cm\)) evaluated under this filtered regime.

Thanks to the iterative nature of our pose hypothesis generation, our framework seamlessly adapts to temporal tracking sequences. We evaluate our approach on the challenging REAL275 dataset benchmark, reporting accuracy metrics alongside tracking speed in Table \ref{tab:track}. We evaluate three distinct variants of our method to highlight different temporal configurations. First, Ours (S) represents our baseline tracking mode; it follows the sequential style of GenPose \cite{zhang_genpose_nodate}, where the final single pose estimate from the previous frame initialises the current frame. Second, Ours (M) denotes the full tracking framework introduced in Section \ref{sec:pose_tracking}, which leverages multi-hypothesis tracking. Finally, Ours (F) represents a non-tracking fallback that discards all temporal context, completely re-initialising the diffusion process from the Gaussian prior at every frame.

As demonstrated in Table \ref{tab:track}, all variants of our method significantly outperform contemporary baselines. Most notably on the strict joint \(10^{\circ}10cm\) metric, our baseline configuration, Ours (S), outperforms GIVEPose by $15\%$ while operating at a highly competitive $6.8\text{ FPS}$. By actively maintaining the previous frame's multi-hypothesis pose distribution, Ours (M) further refines these predictions to achieve a higher joint accuracy of $60.5\%$ at no additional computational cost to the frame rate. This performance bump explicitly emphasises the mathematical benefit of preserving the full tracking distribution to hedge against localised multi-modal ambiguities over time.

When absolute precision is required, Ours (F) establishes a new state-of-the-art baseline on this benchmark at $61.1\%$ joint accuracy—representing a substantial $+18.4\%$ margin over GIVEPose and a $+36.7\%$ improvement over LaPose. This variant underscores the fundamental accuracy-versus-speed trade-off inherent to iterative score-based samplers,where executing full diffusion trajectories maximises alignment precision by comprehensively denoising the pose distribution, albeit at the expense of a reduced frame rate ($3.6\text{ FPS}$). Video examples demonstrating the qualitative stability of these approaches are provided in the Supplementary Material.

\subsection{Ablation Studies}

\subsubsection{Effectiveness of Mean Shift}
To evaluate the isolated impact of our mode-seeking aggregation, we ablate the performance and inference speed of different hypothesis filtering and aggregation configurations on the REAL275 dataset. The results, including frames per second for static single-frame estimation (S-FPS) and temporal tracking (T-FPS), are summarised in Table \ref{tab:rank}.

Incorporating GenPose's \cite{zhang_genpose_nodate} dual-network paradigm — which trains an auxiliary EnergyNet to filter out low-likelihood hypotheses before mean pooling — yields a $46.4$ on the \(10^{\circ}10cm\) metric. However, evaluating raw energy scores through an entirely separate network architecture introduces a severe computational bottleneck, dragging inference speeds down to $3.1$ S-FPS and $3.6$ T-FPS. In contrast, applying our proposed Mean Shift++ aggregation directly to the unfiltered hypothesis set (row 4) completely circumvents this computational penalty. Mean Shift achieves a superior $47.0$ on the \(10^{\circ}10cm\) metric while operating at $3.6$ S-FPS and $6.8$ T-FPS. 

By eliminating the auxiliary filtering network in favour of direct mode-seeking, our pipeline delivers an approximate $15\%$ speed improvement for static pose estimation and a massive $2\times$ speedup for temporal tracking compared to GenPose's baseline.Furthermore, combining EnergyNet filtering with Mean Shift clustering (row 3) establishes the performance upper bound, notably advancing the strict $\text{IoU}_{50}$ to $23.4$. 

This consistent improvement across both filtered and unfiltered configurations confirms that isolating the analytical peak of the empirical distribution is fundamentally more robust than expectation-based mean pooling, successfully bypassing the mean-mode discrepancy inherent to symmetric and multi-modal pose spaces.

\section{Conclusion}
\label{sec:conclusion}

In this work, we introduced a novel framework for RGB-based category-level object pose and size estimation that explicitly addresses multi-hypothesis ambiguity through score-based diffusion modelling. By decoupling feature extraction from absolute 3D coordinate spaces and conditioning our ScoreNet on a joint 2D semantic-geometric context vector, we resolve scale ambiguity without relying on explicit depth sensors. Crucially, we presented a mathematically elegant maximum likelihood formulation that utilises Mean Shift++ to isolate the analytical density peak of the generated hypothesis distribution. This approach completely bypasses the mean-mode discrepancy inherent to traditional mean-based pooling while eliminating the need for expensive auxiliary filtering networks, yielding a highly efficient $2\times$ inference speedup for pose tracking. Our framework establishes a new state-of-the-art on the challenging REAL275 benchmark, outperforming existing RGB baselines by a significant 37\% margin on the strict $10^\circ\text{10cm}$ metric. Finally, we demonstrated that explicitly propagating this multi-modal distribution across sequential frames enables robust temporal tracking in video sequences without architectural adaptations.

While our approach offers substantial gains in both accuracy and latency, generative sampling via the Probability Flow ODE remains a core computational component. Future work will investigate the integration of accelerated diffusion samplers and active hypothesis pruning during the reverse denoising process to further optimise runtime. Additionally, we plan to explore end-to-end joint training paradigms to minimise downstream sensitivity to upstream relative depth and surface normal estimations, paving the way for even more robust unconstrained 3D object understanding.

\bibliographystyle{plain}
\bibliography{egbib}

@String(CVPR= {IEEE Conf. Comput. Vis. Pattern Recog.})

@String(ICCV= {Int. Conf. Comput. Vis.})

@String(ECCV= {Eur. Conf. Comput. Vis.})

@String(CVPR  = {CVPR})

@String(ICCV  = {ICCV})

@String(ECCV  = {ECCV})

@inproceedings{chen_sgpa_2021,
	title = {{SGPA}: Structure-Guided Prior Adaptation for Category-Level 6D Object Pose Estimation},
	doi = {10.1109/ICCV48922.2021.00277},
	shorttitle = {{SGPA}},
	eventtitle = {2021 {IEEE}/{CVF} International Conference on Computer Vision ({ICCV})},
	pages = {2753--2762},
	booktitle = {2021 {IEEE}/{CVF} International Conference on Computer Vision ({ICCV})},
	author = {Chen, Kai and Dou, Qi},
	date = {2021-10},
	note = {{ISSN}: 2380-7504},
}

@inproceedings{fan_object_2022,
	location = {Cham},
	title = {Object Level Depth Reconstruction for Category Level 6D Object Pose Estimation from Monocular {RGB} Image},
	isbn = {978-3-031-20086-1},
	doi = {10.1007/978-3-031-20086-1_13},
	series = {Lecture Notes in Computer Science},
	pages = {220--236},
	booktitle = {Computer Vision – {ECCV} 2022},
	publisher = {Springer Nature Switzerland},
	author = {Fan, Zhaoxin and Song, Zhenbo and Xu, Jian and Wang, Zhicheng and Wu, Kejian and Liu, Hongyan and He, Jun},
	editor = {Avidan, Shai and Brostow, Gabriel and Cissé, Moustapha and Farinella, Giovanni Maria and Hassner, Tal},
	date = {2022},
	langid = {english},
}

@inproceedings{di_gpv-pose_2022,
	location = {New Orleans, {LA}, {USA}},
	title = {{GPV}-Pose: Category-level Object Pose Estimation via Geometry-guided Point-wise Voting},
	isbn = {978-1-66546-946-3},
	url = {https://ieeexplore.ieee.org/document/9880386/},
	doi = {10.1109/CVPR52688.2022.00666},
	shorttitle = {{GPV}-Pose},
	eventtitle = {2022 {IEEE}/{CVF} Conference on Computer Vision and Pattern Recognition ({CVPR})},
	pages = {6771--6781},
	booktitle = {2022 {IEEE}/{CVF} Conference on Computer Vision and Pattern Recognition ({CVPR})},
	publisher = {{IEEE}},
	author = {Di, Yan and Zhang, Ruida and Lou, Zhiqiang and Manhardt, Fabian and Ji, Xiangyang and Navab, Nassir and Tombari, Federico},
	urldate = {2023-07-25},
	date = {2022-06},
	langid = {english},
}

@incollection{zhang_rbp-pose_2022,
	location = {Cham},
	title = {{RBP}-Pose: Residual Bounding Box Projection for Category-Level Pose Estimation},
	volume = {13661},
	isbn = {978-3-031-19768-0 978-3-031-19769-7},
	url = {https://link.springer.com/10.1007/978-3-031-19769-7_38},
	shorttitle = {{RBP}-Pose},
	pages = {655--672},
	booktitle = {Computer Vision – {ECCV} 2022},
	publisher = {Springer Nature Switzerland},
	author = {Zhang, Ruida and Di, Yan and Lou, Zhiqiang and Manhardt, Fabian and Tombari, Federico and Ji, Xiangyang},
	editor = {Avidan, Shai and Brostow, Gabriel and Cissé, Moustapha and Farinella, Giovanni Maria and Hassner, Tal},
	urldate = {2024-02-09},
	date = {2022},
	langid = {english},
	doi = {10.1007/978-3-031-19769-7_38},
	note = {Series Title: Lecture Notes in Computer Science},
}

@article{lee_category-level_2021,
	title = {Category-Level Metric Scale Object Shape and Pose Estimation},
	volume = {6},
	issn = {2377-3766, 2377-3774},
	url = {http://arxiv.org/abs/2109.00326},
	doi = {10.1109/LRA.2021.3110538},
	pages = {8575--8582},
	number = {4},
	journaltitle = {{IEEE} Robotics and Automation Letters},
	shortjournal = {{IEEE} Robot. Autom. Lett.},
	author = {Lee, Taeyeop and Lee, Byeong-Uk and Kim, Myungchul and Kweon, In So},
	urldate = {2024-02-09},
	date = {2021-10},
	langid = {english},
	eprinttype = {arxiv},
	eprint = {2109.00326 [cs]},
	note = {Number: 4},
}

@inproceedings{lin_sar-net_2022,
	location = {New Orleans, {LA}, {USA}},
	title = {{SAR}-Net: Shape Alignment and Recovery Network for Category-level 6D Object Pose and Size Estimation},
	isbn = {978-1-66546-946-3},
	url = {https://ieeexplore.ieee.org/document/9879530/},
	doi = {10.1109/CVPR52688.2022.00659},
	shorttitle = {{SAR}-Net},
	eventtitle = {2022 {IEEE}/{CVF} Conference on Computer Vision and Pattern Recognition ({CVPR})},
	pages = {6697--6707},
	booktitle = {2022 {IEEE}/{CVF} Conference on Computer Vision and Pattern Recognition ({CVPR})},
	publisher = {{IEEE}},
	author = {Lin, Haitao and Liu, Zichang and Cheang, Chilam and Fu, Yanwei and Guo, Guodong and Xue, Xiangyang},
	urldate = {2024-02-09},
	date = {2022-06},
	langid = {english},
}

@article{song_score-based_2021,
	title = {{SCORE}-{BASED} {GENERATIVE} {MODELING} {THROUGH} {STOCHASTIC} {DIFFERENTIAL} {EQUATIONS}},
	author = {Song, Yang and Sohl-Dickstein, Jascha and Kingma, Diederik P and Kumar, Abhishek and Ermon, Stefano and Poole, Ben},
	date = {2021},
	langid = {english},
}

@article{zhang_genpose_nodate,
	title = {{GenPose}: Generative Category-level Object Pose Estimation via Diffusion Models},
	author = {Zhang, Jiyao and Wu, Mingdong and Dong, Hao},
	langid = {english},
}

@inproceedings{wang_normalized_2019,
	location = {Long Beach, {CA}, {USA}},
	title = {Normalized Object Coordinate Space for Category-Level 6D Object Pose and Size Estimation},
	isbn = {978-1-72813-293-8},
	url = {https://ieeexplore.ieee.org/document/8953761/},
	doi = {10.1109/CVPR.2019.00275},
	eventtitle = {2019 {IEEE}/{CVF} Conference on Computer Vision and Pattern Recognition ({CVPR})},
	pages = {2637--2646},
	booktitle = {2019 {IEEE}/{CVF} Conference on Computer Vision and Pattern Recognition ({CVPR})},
	publisher = {{IEEE}},
	author = {Wang, He and Sridhar, Srinath and Huang, Jingwei and Valentin, Julien and Song, Shuran and Guibas, Leonidas J.},
	urldate = {2024-03-06},
	date = {2019-06},
	langid = {english},
}

@inproceedings{chen_category_2020,
	location = {Cham},
	title = {Category Level Object Pose Estimation via Neural Analysis-by-Synthesis},
	isbn = {978-3-030-58574-7},
	doi = {10.1007/978-3-030-58574-7_9},
	series = {Lecture Notes in Computer Science},
	pages = {139--156},
	booktitle = {Computer Vision – {ECCV} 2020},
	publisher = {Springer International Publishing},
	author = {Chen, Xu and Dong, Zijian and Song, Jie and Geiger, Andreas and Hilliges, Otmar},
	editor = {Vedaldi, Andrea and Bischof, Horst and Brox, Thomas and Frahm, Jan-Michael},
	date = {2020},
	langid = {english},
}

@inproceedings{song_generative_2019,
	title = {Generative Modeling by Estimating Gradients of the Data Distribution},
	volume = {32},
	url = {https://proceedings.neurips.cc/paper_files/paper/2019/hash/3001ef257407d5a371a96dcd947c7d93-Abstract.html},
	booktitle = {Advances in Neural Information Processing Systems},
	publisher = {Curran Associates, Inc.},
	author = {Song, Yang and Ermon, Stefano},
	urldate = {2024-03-06},
	date = {2019},
}

@article{manhardt_cps_nodate,
	title = {{CPS}++: Improving Class-level 6D Pose and Shape Estimation From Monocular Images With Self-Supervised Learning},
	author = {Manhardt, Fabian and Wang, Gu and Busam, Benjamin and Nickel, Manuel and Meier, Sven and Minciullo, Luca and Ji, Xiangyang and Navab, Nassir},
	langid = {english},
}

@incollection{tian_shape_2020,
	location = {Cham},
	title = {Shape Prior Deformation for Categorical 6D Object Pose and Size Estimation},
	volume = {12366},
	isbn = {978-3-030-58588-4 978-3-030-58589-1},
	url = {https://link.springer.com/10.1007/978-3-030-58589-1_32},
	pages = {530--546},
	booktitle = {Computer Vision – {ECCV} 2020},
	publisher = {Springer International Publishing},
	author = {Tian, Meng and Ang, Marcelo H. and Lee, Gim Hee},
	editor = {Vedaldi, Andrea and Bischof, Horst and Brox, Thomas and Frahm, Jan-Michael},
	urldate = {2024-03-06},
	date = {2020},
	langid = {english},
	doi = {10.1007/978-3-030-58589-1_32},
	note = {Series Title: Lecture Notes in Computer Science},
}

@article{umeyama_least-squares_1991,
	title = {Least-squares estimation of transformation parameters between two point patterns},
	volume = {13},
	issn = {1939-3539},
	url = {https://ieeexplore.ieee.org/document/88573},
	doi = {10.1109/34.88573},
	pages = {376--380},
	number = {4},
	journaltitle = {{IEEE} Transactions on Pattern Analysis and Machine Intelligence},
	author = {Umeyama, S.},
	urldate = {2024-03-06},
	date = {1991-04},
	note = {Number: 4
Conference Name: {IEEE} Transactions on Pattern Analysis and Machine Intelligence},
}

@inproceedings{wang_gdr-net_2021,
	location = {Nashville, {TN}, {USA}},
	title = {{GDR}-Net: Geometry-Guided Direct Regression Network for Monocular 6D Object Pose Estimation},
	isbn = {978-1-66544-509-2},
	url = {https://ieeexplore.ieee.org/document/9578682/},
	doi = {10.1109/CVPR46437.2021.01634},
	shorttitle = {{GDR}-Net},
	eventtitle = {2021 {IEEE}/{CVF} Conference on Computer Vision and Pattern Recognition ({CVPR})},
	pages = {16606--16616},
	booktitle = {2021 {IEEE}/{CVF} Conference on Computer Vision and Pattern Recognition ({CVPR})},
	publisher = {{IEEE}},
	author = {Wang, Gu and Manhardt, Fabian and Tombari, Federico and Ji, Xiangyang},
	urldate = {2024-03-06},
	date = {2021-06},
	langid = {english},
}

@inproceedings{li_cdpn_2019,
	location = {Seoul, Korea (South)},
	title = {{CDPN}: Coordinates-Based Disentangled Pose Network for Real-Time {RGB}-Based 6-{DoF} Object Pose Estimation},
	isbn = {978-1-72814-803-8},
	url = {https://ieeexplore.ieee.org/document/9009519/},
	doi = {10.1109/ICCV.2019.00777},
	shorttitle = {{CDPN}},
	eventtitle = {2019 {IEEE}/{CVF} International Conference on Computer Vision ({ICCV})},
	pages = {7677--7686},
	booktitle = {2019 {IEEE}/{CVF} International Conference on Computer Vision ({ICCV})},
	publisher = {{IEEE}},
	author = {Li, Zhigang and Wang, Gu and Ji, Xiangyang},
	urldate = {2024-03-06},
	date = {2019-10},
	langid = {english},
}

@inproceedings{zhou_continuity_2019,
	location = {Long Beach, {CA}, {USA}},
	title = {On the Continuity of Rotation Representations in Neural Networks},
	isbn = {978-1-72813-293-8},
	url = {https://ieeexplore.ieee.org/document/8953486/},
	doi = {10.1109/CVPR.2019.00589},
	eventtitle = {2019 {IEEE}/{CVF} Conference on Computer Vision and Pattern Recognition ({CVPR})},
	pages = {5738--5746},
	booktitle = {2019 {IEEE}/{CVF} Conference on Computer Vision and Pattern Recognition ({CVPR})},
	publisher = {{IEEE}},
	author = {Zhou, Yi and Barnes, Connelly and Lu, Jingwan and Yang, Jimei and Li, Hao},
	urldate = {2024-03-06},
	date = {2019-06},
	langid = {english},
}

@inproceedings{he_mask_2017,
	title = {Mask R-{CNN}},
	url = {https://openaccess.thecvf.com/content_iccv_2017/html/He_Mask_R-CNN_ICCV_2017_paper.html},
	eventtitle = {Proceedings of the {IEEE} International Conference on Computer Vision},
	pages = {2961--2969},
	author = {He, Kaiming and Gkioxari, Georgia and Dollar, Piotr and Girshick, Ross},
	urldate = {2024-03-06},
	date = {2017},
}

@inproceedings{wang_category-level_2021,
	title = {Category-Level 6D Object Pose Estimation via Cascaded Relation and Recurrent Reconstruction Networks},
	url = {https://ieeexplore.ieee.org/abstract/document/9636212},
	doi = {10.1109/IROS51168.2021.9636212},
	eventtitle = {2021 {IEEE}/{RSJ} International Conference on Intelligent Robots and Systems ({IROS})},
	pages = {4807--4814},
	booktitle = {2021 {IEEE}/{RSJ} International Conference on Intelligent Robots and Systems ({IROS})},
	author = {Wang, Jiaze and Chen, Kai and Dou, Qi},
	urldate = {2024-03-06},
	date = {2021-09},
	note = {{ISSN}: 2153-0866},
}

@inproceedings{lin_category-level_2022,
	location = {Cham},
	title = {Category-Level 6D Object Pose and Size Estimation Using Self-supervised Deep Prior Deformation Networks},
	isbn = {978-3-031-20077-9},
	doi = {10.1007/978-3-031-20077-9_2},
	series = {Lecture Notes in Computer Science},
	pages = {19--34},
	booktitle = {Computer Vision – {ECCV} 2022},
	publisher = {Springer Nature Switzerland},
	author = {Lin, Jiehong and Wei, Zewei and Ding, Changxing and Jia, Kui},
	editor = {Avidan, Shai and Brostow, Gabriel and Cissé, Moustapha and Farinella, Giovanni Maria and Hassner, Tal},
	date = {2022},
	langid = {english},
}

@inproceedings{zhang_ssp-pose_2022,
	title = {{SSP}-Pose: Symmetry-Aware Shape Prior Deformation for Direct Category-Level Object Pose Estimation},
	url = {https://ieeexplore.ieee.org/abstract/document/9981506},
	doi = {10.1109/IROS47612.2022.9981506},
	shorttitle = {{SSP}-Pose},
	eventtitle = {2022 {IEEE}/{RSJ} International Conference on Intelligent Robots and Systems ({IROS})},
	pages = {7452--7459},
	booktitle = {2022 {IEEE}/{RSJ} International Conference on Intelligent Robots and Systems ({IROS})},
	author = {Zhang, Ruida and Di, Yan and Manhardt, Fabian and Tombari, Federico and Ji, Xiangyang},
	urldate = {2024-03-06},
	date = {2022-10},
	note = {{ISSN}: 2153-0866},
}

@inproceedings{chen_learning_2020,
	title = {Learning Canonical Shape Space for Category-Level 6D Object Pose and Size Estimation},
	url = {https://openaccess.thecvf.com/content_CVPR_2020/html/Chen_Learning_Canonical_Shape_Space_for_Category-Level_6D_Object_Pose_and_CVPR_2020_paper.html},
	eventtitle = {Proceedings of the {IEEE}/{CVF} Conference on Computer Vision and Pattern Recognition},
	pages = {11973--11982},
	author = {Chen, Dengsheng and Li, Jun and Wang, Zheng and Xu, Kai},
	urldate = {2024-03-06},
	date = {2020},
}

@inproceedings{song_denoising_2020,
	title = {Denoising Diffusion Implicit Models},
	url = {https://openreview.net/forum?id=St1giarCHLP},
	eventtitle = {International Conference on Learning Representations},
	author = {Song, Jiaming and Meng, Chenlin and Ermon, Stefano},
	urldate = {2024-03-06},
	date = {2020-10-02},
	langid = {english},
}

@article{vincent_connection_2011,
	title = {A Connection Between Score Matching and Denoising Autoencoders},
	volume = {23},
	issn = {0899-7667},
	url = {https://ieeexplore.ieee.org/document/6795935},
	doi = {10.1162/NECO_a_00142},
	pages = {1661--1674},
	number = {7},
	journaltitle = {Neural Computation},
	author = {Vincent, Pascal},
	urldate = {2024-03-06},
	date = {2011-07},
	note = {Number: 7
Conference Name: Neural Computation},
}

@article{hyvarinen_estimation_nodate,
	title = {Estimation of Non-Normalized Statistical Models by Score Matching},
	author = {Hyvarinen, Aapo},
	langid = {english},
}

@inproceedings{lin_vi-net_2023,
	title = {{VI}-Net: Boosting Category-level 6D Object Pose Estimation via Learning Decoupled Rotations on the Spherical Representations},
	url = {https://openaccess.thecvf.com/content/ICCV2023/html/Lin_VI-Net_Boosting_Category-level_6D_Object_Pose_Estimation_via_Learning_Decoupled_ICCV_2023_paper.html},
	shorttitle = {{VI}-Net},
	eventtitle = {Proceedings of the {IEEE}/{CVF} International Conference on Computer Vision},
	pages = {14001--14011},
	author = {Lin, Jiehong and Wei, Zewei and Zhang, Yabin and Jia, Kui},
	urldate = {2024-03-06},
	date = {2023},
	langid = {english},
}

@incollection{labbe_cosypose_2020,
	location = {Cham},
	title = {{CosyPose}: Consistent Multi-view Multi-object 6D Pose Estimation},
	volume = {12362},
	isbn = {978-3-030-58519-8 978-3-030-58520-4},
	url = {https://link.springer.com/10.1007/978-3-030-58520-4_34},
	shorttitle = {{CosyPose}},
	pages = {574--591},
	booktitle = {Computer Vision – {ECCV} 2020},
	publisher = {Springer International Publishing},
	author = {Labbé, Yann and Carpentier, Justin and Aubry, Mathieu and Sivic, Josef},
	editor = {Vedaldi, Andrea and Bischof, Horst and Brox, Thomas and Frahm, Jan-Michael},
	urldate = {2024-03-07},
	date = {2020},
	langid = {english},
	doi = {10.1007/978-3-030-58520-4_34},
	note = {Series Title: Lecture Notes in Computer Science},
}

@article{li_deepim_nodate,
	title = {{DeepIM}: Deep Iterative Matching for 6D Pose Estimation},
	author = {Li, Yi and Wang, Gu and Ji, Xiangyang and Xiang, Yu and Fox, Dieter},
	langid = {english},
}

@inproceedings{su_zebrapose_2022,
	location = {New Orleans, {LA}, {USA}},
	title = {{ZebraPose}: Coarse to Fine Surface Encoding for 6DoF Object Pose Estimation},
	isbn = {978-1-66546-946-3},
	url = {https://ieeexplore.ieee.org/document/9879853/},
	doi = {10.1109/CVPR52688.2022.00662},
	shorttitle = {{ZebraPose}},
	eventtitle = {2022 {IEEE}/{CVF} Conference on Computer Vision and Pattern Recognition ({CVPR})},
	pages = {6728--6738},
	booktitle = {2022 {IEEE}/{CVF} Conference on Computer Vision and Pattern Recognition ({CVPR})},
	publisher = {{IEEE}},
	author = {Su, Yongzhi and Saleh, Mahdi and Fetzer, Torben and Rambach, Jason and Navab, Nassir and Busam, Benjamin and Stricker, Didier and Tombari, Federico},
	urldate = {2024-03-07},
	date = {2022-06},
	langid = {english},
}

@inproceedings{shugurov_osop_2022,
	location = {New Orleans, {LA}, {USA}},
	title = {{OSOP}: A Multi-Stage One Shot Object Pose Estimation Framework},
	isbn = {978-1-66546-946-3},
	url = {https://ieeexplore.ieee.org/document/9880271/},
	doi = {10.1109/CVPR52688.2022.00671},
	shorttitle = {{OSOP}},
	eventtitle = {2022 {IEEE}/{CVF} Conference on Computer Vision and Pattern Recognition ({CVPR})},
	pages = {6825--6834},
	booktitle = {2022 {IEEE}/{CVF} Conference on Computer Vision and Pattern Recognition ({CVPR})},
	publisher = {{IEEE}},
	author = {Shugurov, Ivan and Li, Fu and Busam, Benjamin and Ilic, Slobodan},
	urldate = {2024-03-07},
	date = {2022-06},
	langid = {english},
}

@article{tang_3d_2020,
	title = {3D Mapping and 6D Pose Computation for Real Time Augmented Reality on Cylindrical Objects},
	volume = {30},
	issn = {1558-2205},
	url = {https://ieeexplore.ieee.org/abstract/document/8887213?casa_token=q8imA1F76jQAAAAA:i_AVIM7GA9qTiHCpNCrCRpFUh33_RKdd9CWTuEVXVHJjEHr4jhNULe-F7x6mVt6sknKq2fYJrnV1},
	doi = {10.1109/TCSVT.2019.2950449},
	pages = {2887--2899},
	number = {9},
	journaltitle = {{IEEE} Transactions on Circuits and Systems for Video Technology},
	author = {Tang, Fulin and Wu, Yihong and Hou, Xiaohui and Ling, Haibin},
	urldate = {2024-03-07},
	date = {2020-09},
	note = {Number: 9
Conference Name: {IEEE} Transactions on Circuits and Systems for Video Technology},
}

@misc{tan_6d_2017,
	title = {6D Object Pose Estimation with Depth Images: A Seamless Approach for Robotic Interaction and Augmented Reality},
	url = {http://arxiv.org/abs/1709.01459},
	doi = {10.48550/arXiv.1709.01459},
	shorttitle = {6D Object Pose Estimation with Depth Images},
	number = {{arXiv}:1709.01459},
	publisher = {{arXiv}},
	author = {Tan, David Joseph and Navab, Nassir and Tombari, Federico},
	urldate = {2024-03-07},
	date = {2017-09-05},
	eprinttype = {arxiv},
	eprint = {1709.01459 [cs]},
	note = {Issue: {arXiv}:1709.01459},
}

@misc{tremblay_deep_2018,
	title = {Deep Object Pose Estimation for Semantic Robotic Grasping of Household Objects},
	url = {http://arxiv.org/abs/1809.10790},
	doi = {10.48550/arXiv.1809.10790},
	number = {{arXiv}:1809.10790},
	publisher = {{arXiv}},
	author = {Tremblay, Jonathan and To, Thang and Sundaralingam, Balakumar and Xiang, Yu and Fox, Dieter and Birchfield, Stan},
	urldate = {2024-03-07},
	date = {2018-09-27},
	eprinttype = {arxiv},
	eprint = {1809.10790 [cs]},
	note = {Issue: {arXiv}:1809.10790},
}

@article{du_vision-based_2021,
	title = {Vision-based robotic grasping from object localization, object pose estimation to grasp estimation for parallel grippers: a review},
	volume = {54},
	issn = {1573-7462},
	url = {https://doi.org/10.1007/s10462-020-09888-5},
	doi = {10.1007/s10462-020-09888-5},
	shorttitle = {Vision-based robotic grasping from object localization, object pose estimation to grasp estimation for parallel grippers},
	pages = {1677--1734},
	number = {3},
	journaltitle = {Artificial Intelligence Review},
	shortjournal = {Artif Intell Rev},
	author = {Du, Guoguang and Wang, Kai and Lian, Shiguo and Zhao, Kaiyong},
	urldate = {2024-03-07},
	date = {2021-03-01},
	langid = {english},
	note = {Number: 3},
}

@article{comaniciu_mean_2002,
	title = {Mean shift: a robust approach toward feature space analysis},
	volume = {24},
	rights = {https://ieeexplore.ieee.org/Xplorehelp/downloads/license-information/{IEEE}.html},
	issn = {01628828},
	url = {http://ieeexplore.ieee.org/document/1000236/},
	doi = {10.1109/34.1000236},
	shorttitle = {Mean shift},
	pages = {603--619},
	number = {5},
	journaltitle = {{IEEE} Transactions on Pattern Analysis and Machine Intelligence},
	shortjournal = {{IEEE} Trans. Pattern Anal. Machine Intell.},
	author = {Comaniciu, D. and Meer, P.},
	urldate = {2024-05-07},
	date = {2002-05},
	langid = {english},
}

@article{yizong_cheng_mean_1995,
	title = {Mean shift, mode seeking, and clustering},
	volume = {17},
	rights = {https://ieeexplore.ieee.org/Xplorehelp/downloads/license-information/{IEEE}.html},
	issn = {01628828},
	url = {http://ieeexplore.ieee.org/document/400568/},
	doi = {10.1109/34.400568},
	pages = {790--799},
	number = {8},
	journaltitle = {{IEEE} Transactions on Pattern Analysis and Machine Intelligence},
	shortjournal = {{IEEE} Trans. Pattern Anal. Machine Intell.},
	author = {{Yizong Cheng}},
	urldate = {2024-05-07},
	date = {1995-08},
	langid = {english},
}

@misc{lin_microsoft_2015,
	title = {Microsoft {COCO}: Common Objects in Context},
	url = {http://arxiv.org/abs/1405.0312},
	shorttitle = {Microsoft {COCO}},
	number = {{arXiv}:1405.0312},
	publisher = {{arXiv}},
	author = {Lin, Tsung-Yi and Maire, Michael and Belongie, Serge and Bourdev, Lubomir and Girshick, Ross and Hays, James and Perona, Pietro and Ramanan, Deva and Zitnick, C. Lawrence and Dollár, Piotr},
	urldate = {2024-05-22},
	date = {2015-02-20},
	langid = {english},
	eprinttype = {arxiv},
	eprint = {1405.0312 [cs]},
}

@misc{piccinelli_unidepth_2024,
	title = {{UniDepth}: Universal Monocular Metric Depth Estimation},
	url = {http://arxiv.org/abs/2403.18913},
	doi = {10.48550/arXiv.2403.18913},
	shorttitle = {{UniDepth}},
	number = {{arXiv}:2403.18913},
	publisher = {{arXiv}},
	author = {Piccinelli, Luigi and Yang, Yung-Hsu and Sakaridis, Christos and Segu, Mattia and Li, Siyuan and Van Gool, Luc and Yu, Fisher},
	urldate = {2024-05-22},
	date = {2024-03-27},
	eprinttype = {arxiv},
	eprint = {2403.18913 [cs]},
}

@article{kar_3d_nodate,
	title = {3D Common Corruptions and Data Augmentation},
	author = {Kar, Oguzhan Fatih and Yeo, Teresa and Atanov, Andrei and Zamir, Amir},
	langid = {english},
}

@inproceedings{eftekhar_omnidata_2021,
	location = {Montreal, {QC}, Canada},
	title = {Omnidata: A Scalable Pipeline for Making Multi-Task Mid-Level Vision Datasets from 3D Scans},
	rights = {https://doi.org/10.15223/policy-029},
	isbn = {978-1-66542-812-5},
	url = {https://ieeexplore.ieee.org/document/9710360/},
	doi = {10.1109/ICCV48922.2021.01061},
	shorttitle = {Omnidata},
	eventtitle = {2021 {IEEE}/{CVF} International Conference on Computer Vision ({ICCV})},
	pages = {10766--10776},
	booktitle = {2021 {IEEE}/{CVF} International Conference on Computer Vision ({ICCV})},
	publisher = {{IEEE}},
	author = {Eftekhar, Ainaz and Sax, Alexander and Malik, Jitendra and Zamir, Amir},
	urldate = {2024-05-22},
	date = {2021-10},
	langid = {english},
}

@inproceedings{jang_meanshift_2021,
	location = {Nashville, {TN}, {USA}},
	title = {{MeanShift}++: Extremely Fast Mode-Seeking With Applications to Segmentation and Object Tracking},
	rights = {https://ieeexplore.ieee.org/Xplorehelp/downloads/license-information/{IEEE}.html},
	isbn = {978-1-66544-509-2},
	url = {https://ieeexplore.ieee.org/document/9578266/},
	doi = {10.1109/CVPR46437.2021.00409},
	shorttitle = {{MeanShift}++},
	eventtitle = {2021 {IEEE}/{CVF} Conference on Computer Vision and Pattern Recognition ({CVPR})},
	pages = {4100--4111},
	booktitle = {2021 {IEEE}/{CVF} Conference on Computer Vision and Pattern Recognition ({CVPR})},
	publisher = {{IEEE}},
	author = {Jang, Jennifer and Jiang, Heinrich},
	urldate = {2024-05-22},
	date = {2021-06},
	langid = {english},
}

@misc{wei_rgb-based_2023,
	title = {{RGB}-based Category-level Object Pose Estimation via Decoupled Metric Scale Recovery},
	url = {http://arxiv.org/abs/2309.10255},
	number = {{arXiv}:2309.10255},
	publisher = {{arXiv}},
	author = {Wei, Jiaxin and Song, Xibin and Liu, Weizhe and Kneip, Laurent and Li, Hongdong and Ji, Pan},
	urldate = {2024-06-19},
	date = {2023-10-18},
	langid = {english},
	eprinttype = {arxiv},
	eprint = {2309.10255 [cs]},
}

@misc{russakovsky_imagenet_2015,
	title = {{ImageNet} Large Scale Visual Recognition Challenge},
	url = {http://arxiv.org/abs/1409.0575},
	doi = {10.48550/arXiv.1409.0575},
	number = {{arXiv}:1409.0575},
	publisher = {{arXiv}},
	author = {Russakovsky, Olga and Deng, Jia and Su, Hao and Krause, Jonathan and Satheesh, Sanjeev and Ma, Sean and Huang, Zhiheng and Karpathy, Andrej and Khosla, Aditya and Bernstein, Michael and Berg, Alexander C. and Fei-Fei, Li},
	urldate = {2024-12-12},
	date = {2015-01-30},
	langid = {english},
	eprinttype = {arxiv},
	eprint = {1409.0575 [cs]},
}

@misc{he_deep_2015,
	title = {Deep Residual Learning for Image Recognition},
	url = {http://arxiv.org/abs/1512.03385},
	doi = {10.48550/arXiv.1512.03385},
	number = {{arXiv}:1512.03385},
	publisher = {{arXiv}},
	author = {He, Kaiming and Zhang, Xiangyu and Ren, Shaoqing and Sun, Jian},
	urldate = {2024-12-12},
	date = {2015-12-10},
	langid = {english},
	eprinttype = {arxiv},
	eprint = {1512.03385 [cs]},
}

@misc{huang_givepose_2025,
	title = {{GIVEPose}: Gradual Intra-class Variation Elimination for {RGB}-based Category-Level Object Pose Estimation},
	url = {http://arxiv.org/abs/2503.15110},
	doi = {10.48550/arXiv.2503.15110},
	shorttitle = {{GIVEPose}},
	number = {{arXiv}:2503.15110},
	publisher = {{arXiv}},
	author = {Huang, Zinqin and Wang, Gu and Zhang, Chenyangguang and Zhang, Ruida and Li, Xiu and Ji, Xiangyang},
	urldate = {2025-05-16},
	date = {2025-03-20},
	langid = {english},
	eprinttype = {arxiv},
	eprint = {2503.15110 [cs]},
}

@incollection{leonardis_lapose_2025,
	location = {Cham},
	title = {{LaPose}: Laplacian Mixture Shape Modeling for {RGB}-Based Category-Level Object Pose Estimation},
	volume = {15083},
	isbn = {978-3-031-72697-2 978-3-031-72698-9},
	url = {https://link.springer.com/10.1007/978-3-031-72698-9_27},
	shorttitle = {{LaPose}},
	pages = {467--484},
	booktitle = {Computer Vision – {ECCV} 2024},
	publisher = {Springer Nature Switzerland},
	author = {Zhang, Ruida and Huang, Ziqin and Wang, Gu and Zhang, Chenyangguang and Di, Yan and Zuo, Xingxing and Tang, Jiwen and Ji, Xiangyang},
	editor = {Leonardis, Aleš and Ricci, Elisa and Roth, Stefan and Russakovsky, Olga and Sattler, Torsten and Varol, Gül},
	urldate = {2025-05-16},
	date = {2025},
	langid = {english},
	doi = {10.1007/978-3-031-72698-9_27},
	note = {Series Title: Lecture Notes in Computer Science},
}

@inproceedings{lee2026yopo,
  title     = {You Only Pose Once: A Minimalist’s Detection Transformer for Monocular RGB Category-level 9D Multi-Object Pose Estimation},
  author    = {Lee, Hakjin and Seo, Junghoon and Sim, Jaehoon},
  booktitle = {IEEE International Conference on Robotics and Automation (ICRA)},
  year      = {2026}
}

@InProceedings{MonoDiff9D,
  title={MonoDiff9D: Monocular Category-Level 9D Object Pose Estimation via Diffusion Model},
  author={Liu, Jian and Sun, Wei and Yang, Hui and Zheng, Jin and Geng, Zichen and Rahmani, Hossein and Mian, Ajmal},
  booktitle = {IEEE International Conference on Robotics and Automation (ICRA)},
  year={2025}
}

@misc{fischer2025unifiedcategorylevelobjectdetection,
      title={Unified Category-Level Object Detection and Pose Estimation from RGB Images using 3D Prototypes}, 
      author={Tom Fischer and Xiaojie Zhang and Eddy Ilg},
      year={2025},
      eprint={2508.02157},
      archivePrefix={arXiv},
      primaryClass={cs.CV},
      url={https://arxiv.org/abs/2508.02157}, 
}

@article{LanCope,
  author={Yang, Hui and Sun, Wei and Liu, Jian and Zheng, Jin and Dai, Zhenqi and Mian, Ajmal},
  journal={IEEE Robotics and Automation Letters}, 
  title={LanCOPE: Language-Guided Category-Level Object Pose Estimation From a Single RGB Image}, 
  year={2025},
  volume={10},
  number={7},
  pages={7555-7562},
  doi={10.1109/LRA.2025.3577460}}

@INPROCEEDINGS{rcgnet,
  author={Yu, Sheng and Zhai, Di-Hua and Xia, Yuanqing},
  booktitle={2025 IEEE/RSJ International Conference on Intelligent Robots and Systems (IROS)}, 
  title={RCGNet: RGB-based Category-Level 6D Object Pose Estimation with Geometric Guidance}, 
  year={2025},
  volume={},
  number={},
  pages={740-745},
  doi={10.1109/IROS60139.2025.11247442}}

@article{Diff9D,
  author={Liu, Jian and Sun, Wei and Yang, Hui and Deng, Pengchao and Liu, Chongpei and Sebe, Nicu and Rahmani, Hossein and Mian, Ajmal},
  journal={IEEE Transactions on Pattern Analysis and Machine Intelligence},
  title={Diff9D: Diffusion-Based Domain-Generalized Category-Level 9-DoF Object Pose Estimation},
  year={2025},
  volume={47},
  number={7},
  pages={5520-5537},
  doi={10.1109/TPAMI.2025.3552132}
}

\newpage
\appendix

\section{Architecture and Additional Implementation Details}

\begin{figure*}[h]
  \centering
    \includegraphics[height=7cm]{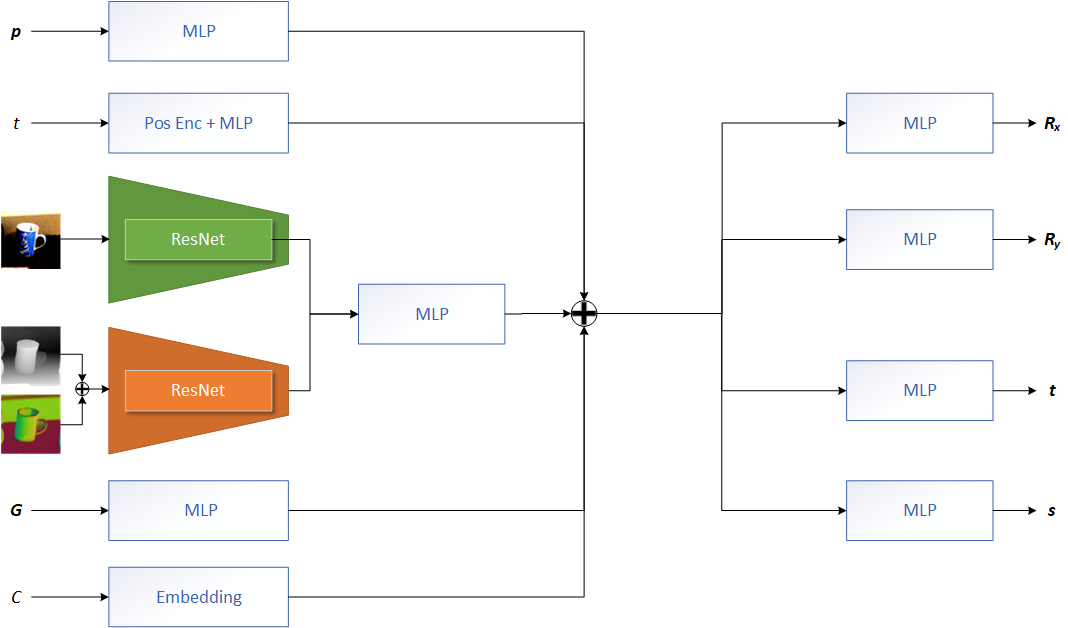}  
  \caption{The architecture of our ScoreNet.}
  \label{fig:arch}
\end{figure*}

\subsection{Implementation Details}
\label{subsec:implementation}
The predicted relative depth and normals for each image are estimated by DPT models pre-trained on Omnidata \cite{eftekhar_omnidata_2021, kar_3d_nodate}. We generate the image patches using the results from MaskRCNN \cite{he_mask_2017} as in previous works to enable a fair comparison. Each patch is resized to $256 \times 256$ before being fed into the ScoreNet. The CNN encoders are ResNet-34s \cite{he_deep_2015} with the image encoder's weights initialised from a pre-trained classification model trained on ImageNet1K \cite{russakovsky_imagenet_2015}. The pose encoder is implemented as an MLP, and time is encoded using the Gaussian Fourier projection from \cite{song_score-based_2021} with a single MLP layer. The rotation, translation, and size heads are MLPs with ReLU activations. All training and testing experiments were conducted on an NVIDIA RTX 4080 GPU with a batch size of 64.

\subsection{Architecture}
The architecture of our ScoreNet is shown in Figure \ref{fig:arch}. It comprises of two ResNet-34s \cite{he_deep_2015} initialised from a pre-trained image classification model trained on ImageNet1K \cite{russakovsky_imagenet_2015}, fours feature encoders and four heads. The inputs to the CNNs are the cropped image and predicted depth and normals estimated by the DPT models pre-trained on Omnidata \cite{eftekhar_omnidata_2021, kar_3d_nodate}. We generate the image patches using the results from MaskRCNN \cite{he_mask_2017} as in previous works to enable a fair comparison. Each patch is resized to \(256 \times 256\) before being fed into the ScoreNet. The pose encoder is implemented as a 2-layer MLP with hidden dimensions 256 while the category Id is a learned embedding of size 128. The global features is passed through an MLP resulting in a 2048 dimensional tensor and the time is encoded using the Gaussian Fourier projection from \cite{song_denoising_2020} with a single MLP layer. The final CNN output is concatenated with all of the encoded features and passed into the four heads. The rotation, translation and size heads are 3-layer MLPs with hidden dimensions of 512 again with ReLU activation in between all the layers.

\subsection{Model Flexibility}
\label{sec:flex}
In order to make our model more flexible and allow the model to be used with different object detectors and for novel categories without retraining, we take inspiration from classifier-free guidance and drop the conditions during training. Specifically, when a condition is dropped during training it is replaced by a null value \(\emptyset\) or a tensor of the same size but filled with zeros. In the case of the category IDs we utilise the zeroth ID as an unknown category. The conditions are dropped randomly at a 10\% rate.

To further improve the flexibility of the model we follow previous object pose estimation methods and decouple the object detection and pose estimation by generating the image crop for the image, depth and normals using Dynamic Zoom In (DZI) \cite{li_cdpn_2019, wang_gdr-net_2021}. This process enables the method to work on top of any object detector and increases the robustness to detection errors.

\subsubsection{Dynamic Zoom In}
\label{sec:dzi}
The input to our CNN encoders are cropped images of the object so following previous object pose estimation methods, we decouple the object detection and pose estimation by generating the image crop for the image, depth and normals using Dynamic Zoom In (DZI) \cite{li_cdpn_2019, wang_gdr-net_2021}. This process enables the method to work on top of any object detector and increases the robustness to detection errors. Specifically when given an image with ground truth object bounding box centre, \((O_x, O_y)\), and size, \(s = max(h,w)\), where \(h\) and \(w\) are the height and width of the bounding box in pixels, we shift \((O_x, O_y)\) uniformly by a ratio of 25\% and scale the \(s\) by the same ratio. A zoom-in ratio of 1.5 is then used to maintain the aspect ratio and ensure the area containing the object is approximately half the cropped image. At inference time, the detected bounding box is zoomed in with the same ratio (1.5) as in training with no additional shifting or scaling.  
\subsubsection{Pose Parameterisation}
\label{sec:pose}
In terms of the pose parameterisation, we parameterise the pose, \(\mathbf{p}\), using a 12D vector with 6D rotation, 3D translation and 3D size components,  \(\mathbf{p} = [\mathbf{R}|\mathbf{T}|\mathbf{s}], \mathbf{R} \in \mathbb{R}^6\), \(\mathbf{T} \in \mathbb{R}^3\) and \(\mathbf{s} \in \mathbb{R}^3\). Specifically, the rotation is the continuous 6D representation \([\mathbf{R}_x|\mathbf{R}_y]\) following \cite{zhou_continuity_2019, wang_gdr-net_2021} due to the discontinuity of Euler angles and quaternions. The 6D representation of the rotation can be converted to \(SO(3)\) using the following equations:

\begin{equation}
    g([\mathbf{a}_1|\mathbf{a}_2]) = [\mathbf{b}_1|\mathbf{b}_2|\mathbf{b}_3]
  \label{eq:6dso3}
\end{equation}

\begin{equation}
 \begin{cases}
      \mathbf{b}_1 = N(\mathbf{a}_1)\\
      \mathbf{b}_2 = N(\mathbf{a}_2 - (\mathbf{b}_1 \cdot\mathbf{a}_2)\mathbf{b}_1)\\
      \mathbf{b}_3 = \mathbf{b}_1 \times \mathbf{b}_2
    \end{cases}
  \label{eq:bs}
\end{equation}

Where \(N(\cdot)\) is the normalisation function. 

The mapping from \(SO(3)\) to the 6D representation is defined as:

\begin{equation}
      f([\mathbf{a}_1|\mathbf{a}_2|\mathbf{a}_3]) = [\mathbf{a}_1|\mathbf{a}_2]
  \label{eq:so36d}
\end{equation}

As with other works in RGB-based object pose estimation with cropped images we utilise the Scale Invariant Translation Estimation (SITE) as the translation parameterisation rather than directly predicting \(\mathbf{T} = [T_x, T_y, T_z]\). In this parameterization, the translation is represented by \((\delta_x, \delta_y, t_z)\) where \(\delta_x\) and \(\delta_y\) represent the offset between the crop centre to the object centre and \(t_z\) is the zoomed-in depth. These can be calculated from the bounding box information and the ground truth object centre using Equation \ref{eq:site}. The relative offset between the crop and object centres is utilised as this is constant with respect to the Dynamic Zoom In. The translation can be solved using the global information of the image patch, the predicted offset and the zoomed-in depth using Equation \ref{eq:trans}. 

\begin{equation}
    \begin{cases}
      \delta_x = \frac{O_x - C_x}{w}\\
      \delta_y = \frac{O_y - C_y}{h}\\
      t_z = \frac{T_z}{r}
    \end{cases}
  \label{eq:site}
\end{equation}

\begin{equation}
 \begin{cases}
      T_x = (\delta_x \cdot w + C_x) \cdot \frac{T_z}{f_x}\\
      T_y = (\delta_y \cdot h + C_y) \cdot \frac{T_z}{f_y}\\
      T_z = r \cdot t_z
    \end{cases}
  \label{eq:trans}
\end{equation}

where \((O_x, O_y)\) and \((C_x, C_y)\) is the projection of the object centre and centre of the cropped image in the original image, \((h,w)\) is the size of the bounding box in the original image and \(r = s_{zoom} / s_o\), \(s_{zoom}\) is the size of the resize image patch and \(s_o = \max(h,w)\).  

Unlike previous methods, we utilise a SITE inspired scaling of the 3D size of the object. Specifically the size of the object \(\mathbf{S} \in \mathbb{R}^3\) is predicted using equation \ref{eq:size}:
\begin{equation}
      S = r \cdot s
  \label{eq:size}
\end{equation}

where \(s\) is the estimated scaled size of the object.

\subsection{Implementation Details}

The ScoreNet is trained for 20 epochs using the Adam optimiser with a base learning rate of 2e-4. During training, the loss components for the rotation, translation, and 3D size regression heads are weighted equally ($\lambda_{rot} = \lambda_{trans} = \lambda_{size} = 1$). To ensure stable generative sampling, we apply an Exponential Moving Average (EMA) to the network weights with a decay rate of 0.999.

Our generative framework is modelled continuously via a Variance Exploding (VE) Stochastic Differential Equation (SDE) bounded between $\sigma_{min} = 0.01$ and $\sigma_{max} = 50$, which fundamentally eliminates the need for discrete training time steps. During inference, we sample $K = 50$ pose hypotheses per object. The reverse Probability Flow ODE is initialised with a maximum noise scale of 0.55 and solved dynamically using the RK45 method, with both relative and absolute error tolerances (rtol and atol) strictly set to 1e-5.

For the final pose aggregation, the Mean Shift++ mode-seeking algorithm converges automatically based on spatial density without requiring a fixed iteration limit. It operates using clustering bandwidths of 0.2 for rotation, 0.05 for translation, and 0.01 for 3D size. The optimal selections for $K$, the initial diffusion noise scale, and the Mean Shift bandwidths are explicitly ablated in Section 4.

\section{Object Pose Tracking Details}
\label{sec:track}
Due to the iterative nature of our pose estimation method, it is able to be used in a pose-tracking framework by utilising the pose hypotheses from the previous frame as the initialisation of the current frame. The full algorithm is outlined below:

\begin{algorithm}
\caption{Pose Tracking Framework}
\label{alg:track}
\begin{algorithmic}
\Require ScoreNet $s_{\theta}$, initial ground truth pose $\mathbf{p}_0$ and tracking horizon $T$.
\For{t=1 \textbf{to} $T$}
    \State Receive current observation \(I_t\)
    \State \(\mathbf{z}_1, \mathbf{z}_2, ..., \mathbf{z}_K \sim \mathcal{N}({\mathbf{p}}_{t-1}^K + \mathbf{\hat{v}_{t-1}}, 0.1^2I)\) \Comment{Add noise to previous pose hypotheses}
    \State Sample new pose hypotheses \({\hat{\mathbf{p}}}^K_{i=1}\) from \({\mathbf{z}_1}^K_{i=1}\)
    \State Estimate current pose \(\mathbf{\hat{p}_t}\) using Mean Shift
    \State Calculate velocity \(\mathbf{\hat{v}_t} = \mathbf{\hat{p}_t} - \mathbf{\hat{p}_{t-1}}\)
\EndFor
\end{algorithmic}
\end{algorithm}

Video examples some of the scenes are provided in the rest of the Supplementary Material which demonstrates our trackers ability to maintain the track of an objects pose by also recover after missed detections or estimation errors.

\begin{table*}[tb]
\begin{center}
  \begin{tabular}{|P{5.5em} | P{2em} P{2em} P{2em} P{2em} P{4em}|}
    \toprule
    \multirow{2}{*}{Method} &\multicolumn{5}{c|}{REAL275} \\
    & IoU25 & IoU50 & \(10^o\) & \(10cm\) & \(10^o10cm\)  \\
    \midrule
    Ours - NCG & 43.9 & 14.8 & 60.4 & 48.0 & 39.9\\
    Ours - NG & 43.3 & 17.1  & 55.7 & 44.3 & 36.8\\
    Ours - NC & 45.4 & 18.6 & 59.7 & 51.6 & 44.4\\
    Ours - All & 46.2 & 19.6 & 55.8 & 54.1 & 42.1\\
    Ours &   \textbf{47.0} &\textbf{22.9}   & \textbf{61.1} & \textbf{54.6} & \textbf{47.0}\\
\bottomrule
  \end{tabular}
  \end{center}
   \caption{Quantitative comparison of different inputs on the REAL275 dataset.}
   \label{tab:input}
\end{table*}

\section{Additional Ablation Study}
\subsection{Ablation Study on Conditioning Inputs}
\label{sec:ablation_inputs}

We conduct an ablation study on the REAL275 dataset to validate our design choices regarding the injection of conditioning inputs—specifically, the object category ID used for classifier-free guidance and the global scene features. Our baseline variations are defined as follows:
\begin{itemize}
    \item \textbf{Ours - NCG:} Omits both the category ID and the global features entirely.
    \item \textbf{Ours - NC:} Retains global features but omits the category ID.
    \item \textbf{Ours - NG:} Retains the category ID but omits global features from all architectural prediction heads.
    \item \textbf{Ours - All:} Injects both the category ID and global features globally across all prediction heads.
    \item \textbf{Ours:} Our final configuration, which injects the category ID across all heads but removes the global features \textit{specifically} from the rotation heads.
\end{itemize}

The quantitative results are summarised in Table~\ref{tab:input}.

\subsubsection{Analysis of Results}

\paragraph{The Impact of Global Context on Translation vs. Rotation.}
A key insight from Table~\ref{tab:input} is the conflicting impact of global features on distinct pose components. Comparing \textbf{Ours - NG} (no global features anywhere) to \textbf{Ours - All} (global features everywhere) reveals that global context drastically improves translation ($10\text{cm}$ jumps from $44.3$ to $54.1$) and 3D bounding box overlap ($\text{IoU50}$ improves from $17.1\%$ to $19.6\%$). This is mathematically intuitive: absolute 3D translation $\mathbf{t}$ and scale $\mathbf{s}$ from a single RGB image are highly dependent on global perspective cues, camera intrinsics, and relative scene depth some of which are lost during cropping. 

However, introducing global features everywhere severely penalises rotation accuracy ($10^\circ$ drops from $60.4$ in NCG to $55.8\%$ in All). We attribute this to feature dilution; global scene embeddings introduce geometric noise that distracts the rotation heads from focusing on local, object-specific orientation cues. By isolating this effect, our final architecture (\textbf{Ours}) selectively removes global features \textit{only} from the rotation heads while retaining them for translation and scale. This hybrid approach achieves the ``best of both worlds,'' maximising translation accuracy ($54.6$) while reclaiming the highest rotation performance ($61.1$).

\paragraph{The Necessity of Category Conditioning for Classifier-Free Guidance.}
Omitting the category ID consistently degrades performance across all spatial metrics. Comparing \textbf{Ours - NC} to \textbf{Ours} demonstrates that the inclusion of the category ID yields a significant boost across the board, particularly visible in the strict $\text{IoU50}$ metric (improving from $18.6$ to $22.9$) and the combined $10^\circ, 10\text{cm}$ metric (improving from $44.4$ to $47.0$). Because category-level pose estimation requires the network to map highly diverse intra-class geometries to a unified canonical space, the category ID acts as a vital prior.

\paragraph{Diffusion Noise Schedule}
We ablated the maximum continuous time variable $T$ ($t \in [0, T]$) used to initialise the reverse Probability Flow ODE. Because our framework uses an adaptive ODE solver, the reverse denoising steps dynamically scale based on the initial noise level and the complexity of the denoising trajectory, directly impacting runtime latency. 

\begin{itemize}
    \item \textbf{Under-Exploration ($T = 0.3$):} Restricts the multi-hypothesis exploration space. The adaptive solver executes too few refinement steps, causing a severe performance drop to 6.3 on $10^{\circ}$10cm while operating at 3.6 FPS.
    \item \textbf{Structural Corruption ($T = 0.8$):} Aggressively corrupts the spatial structure, degrading conditioning alignment and reducing accuracy to 42.4 while dropping inference speed to 2.0 FPS due to the increased trajectory complexity required by the adaptive solver.
    \item \textbf{Optimal Balance ($T = 0.55$):} Provides a sufficient noise horizon to comprehensively refine the pose hypotheses while ensuring stable mode-seeking convergence via Mean Shift, peaking at \textbf{47.0} on $10^{\circ}$10cm at 3.6 FPS (Table \ref{tab:noise}).
\end{itemize}

\subsection{Generative Parameter Sweeps}
Table \ref{tab:hypoth} demonstrates the impact of the sampled hypothesis count ($K$) on both accuracy and inference throughput. At the extreme low end ($K=1$), performance is bottlenecked at 38.3 joint accuracy, though it runs at a faster 9.1 FPS. As the hypothesis count increases, accuracy improves steadily while frame rates decrease. Performance heavily saturates around \textbf{$K=50$} (yielding an optimal joint accuracy of 47.0 at 3.6 FPS). Doubling the hypothesis count to $K=100$ yields only a negligible 0.1 gain in joint accuracy while nearly halving the throughput to 2.1 FPS, making $K=50$ our selected optimal speed-accuracy trade-off.

\subsection{Mean Shift Bandwidth Selection}
Table \ref{tab:mean} details the model's sensitivity to Mean Shift bandwidth parameters. The most stable mode-seeking behaviour is achieved with the following bandwidths:
\begin{itemize}
    \item \textbf{Rotation:} 0.2 (corresponding to $\approx 10^{\circ}$)
    \item \textbf{Translation:} 0.05 (corresponding to 5cm)
    \item \textbf{Scale:} 0.01 (corresponding to 1cm)
\end{itemize}

\subsection{GenPose with Predicted Metric Depth}
To validate our structural departures from the original GenPose architecture, we evaluated a baseline naively adapting GenPose to an RGB-only setting by swapping its ground-truth point cloud with a 3D point cloud back-projected from predicted metric depths (Table \ref{tab:metric}).

\begin{itemize}
    \item \textbf{Metric Depth Sensitivity (UniDepth):} Feeding back-projected metric depths from a state-of-the-art estimator like UniDepth into GenPose yields inferior performance (18.2 on $10^\circ$10cm). GenPose's 3D PointNet++ backbone is highly sensitive to global scale drift; scale errors warp spatial coordinates outside the network's rigid metric grouping radii, severely degrading feature representations. 
    \item \textbf{Relative Depth Collapse:} Substituting metric predictions with relative depth maps inside the unmodified GenPose pipeline causes total collapse (0.8 on $10^\circ$10cm). A point cloud coordinate backbone lacks the capacity to infer absolute metrics without spatial scaling bounds.
    \item \textbf{Our Decoupled Advantage:} By processing 2D relative depth ($I_D$) and surface normals ($I_n$) through image backbones—and recovering scale downstream via category embeddings and global image features—our architecture bypasses 3D coordinate tensor warping entirely, achieving \textbf{47.0}.
\end{itemize}

\begin{table*}[tb]
  \centering
  \begin{tabular}{|P{4em}| P{2em} P{2em} P{2em} P{2em}  P{5em}| P{3em}|}
    \toprule
    Noise & IoU25 & IoU50 & \(10^o\) & \(10cm\) & \(10^o10cm\) & FPS\\
    \midrule
    0.3 &  1.6 & 0.1 & 18.8 & 7.3 & 6.3 & 3.6 \\
    0.4 &  41.8 & 12.4 & 60.1 & 48.2 & 39.0 & 4.6 \\
    0.5 &  \textbf{49.6} & 19.8 & \textbf{62.9} & \textbf{54.6} & 46.1 & 3.8 \\
    0.55 &  47.0 & \textbf{22.9} & 61.3 & \textbf{54.6} & \textbf{47.0} & 3.6 \\
    0.6 &  44.5 & 20.3 & 59.8 & 52.6 & 45.4 & 3.3 \\ 
    0.7 &  42.1 & 19.9 & 58.5 & 50.2 & 43.6 & 2.7 \\ 
    0.8 & 41.2 & 19.5 & 57.8 & 48.7 & 42.4 & 2.0 \\
  \bottomrule
  \end{tabular}
  \caption{Ablation on the initial diffusion timestep $T$ on the REAL275 dataset.}
  \label{tab:noise}
\end{table*}

\begin{table*}[tb]
  \centering
  \begin{tabular}{|P{4em}| P{2em} P{2em} P{2em} P{2em}  P{5em}| P{3em}|}
    \toprule
    K & IoU25 & IoU50 & \(10^o\) & \(10cm\) & \(10^o10cm\) & FPS \\
    \midrule
    1  &  41.9 & 18.5 & 54.3 & 49.6 & 38.3 & 9.1 \\
    5  & 44.4 & 20.1 & 58.9 & 52.9 & 44.1 & 8.2 \\
    10 &  46.0 & 21.0 & 60.1 & 53.8 & 45.6 & 7.5 \\
    25 & 46.3 & 21.9 & 60.7 & 53.9 & 46.1 & 5.2\\
    50 &  47.0 & 22.9 & 61.3 & 54.6 & 47.0 & 3.6 \\
    100 &  \textbf{47.1} &\textbf{23.0} & \textbf{61.5} & \textbf{54.7} & \textbf{47.1} & 2.1 \\
  \bottomrule
  \end{tabular}
  \caption{Ablation on the number of hypothesis on the REAL275 dataset.}
  \label{tab:hypoth}
\end{table*}

\begin{table*}[tb]
  \centering
  \begin{tabular}{|P{4em} P{4em} P{4em}| P{2em} P{2em} P{2em} P{2em}  P{5em}|}
    \toprule
    Rotation & Translation & Scale & IoU25 & IoU50 & \(10^o\) & \(10cm\) & \(10^o10cm\) \\
    \midrule
    0.1  &  0.05 & 0.01 & 46.6 & 22.4 & \textbf{61.3} & 54.3 & 46.9  \\
    0.2  & 0.1 & 0.01 & 46.6 & 22.3 & \textbf{61.3} & 54.3 & 46.8  \\
    0.3 & 0.05 & 0.01 & 46.5 & 22.4 & \textbf{61.3} & 54.3 & 46.8 \\
    0.2 & 0.02 & 0.01 & 46.4 & 22.0 & 61.0 & 53.9 & 46.1 \\
    0.2 & 0.05 & 0.02 & 46.6 & 22.5 & \textbf{61.3} & 54.3 & 46.8 \\
    0.2 & 0.05 & 0.005 & 46.2 & 22.0 & 61.2 & 54.4 & 46.8 \\
    0.2 & 0.05 & 0.01 & \textbf{47.0} &\textbf{22.9} & \textbf{61.3} & \textbf{54.6} & \textbf{47.0} \\
  \bottomrule
  \end{tabular}
  \caption{Ablation on the Mean Shift bandwidths on the REAL275 dataset.}
  \label{tab:mean}
\end{table*}

\begin{table*}[h]
  \centering
  \begin{tabular}{|P{10em} | P{3em} P{3em} P{4em}|}
    \toprule
    Method &  \(10^o\) & \(10cm\) & \(10^o10cm\)\\
    \midrule
    GenPose w/ Rel Depth & 16.9 & 9.7 & 0.8 \\
    GenPose w/ UniDepth   & 33.6 & 29.3 & 18.2  \\
    Ours  & \textbf{61.3} & \textbf{54.6} & \textbf{47.0}  \\
  \bottomrule
  \end{tabular}
  \caption{Quantitative results of predicted metric depth in GenPose on the REAL275 dataset.}
  \label{tab:metric}
\end{table*}

\section{Per Category Results}
\begin{figure}[tb]
  \centering
  \includegraphics[height=6cm]{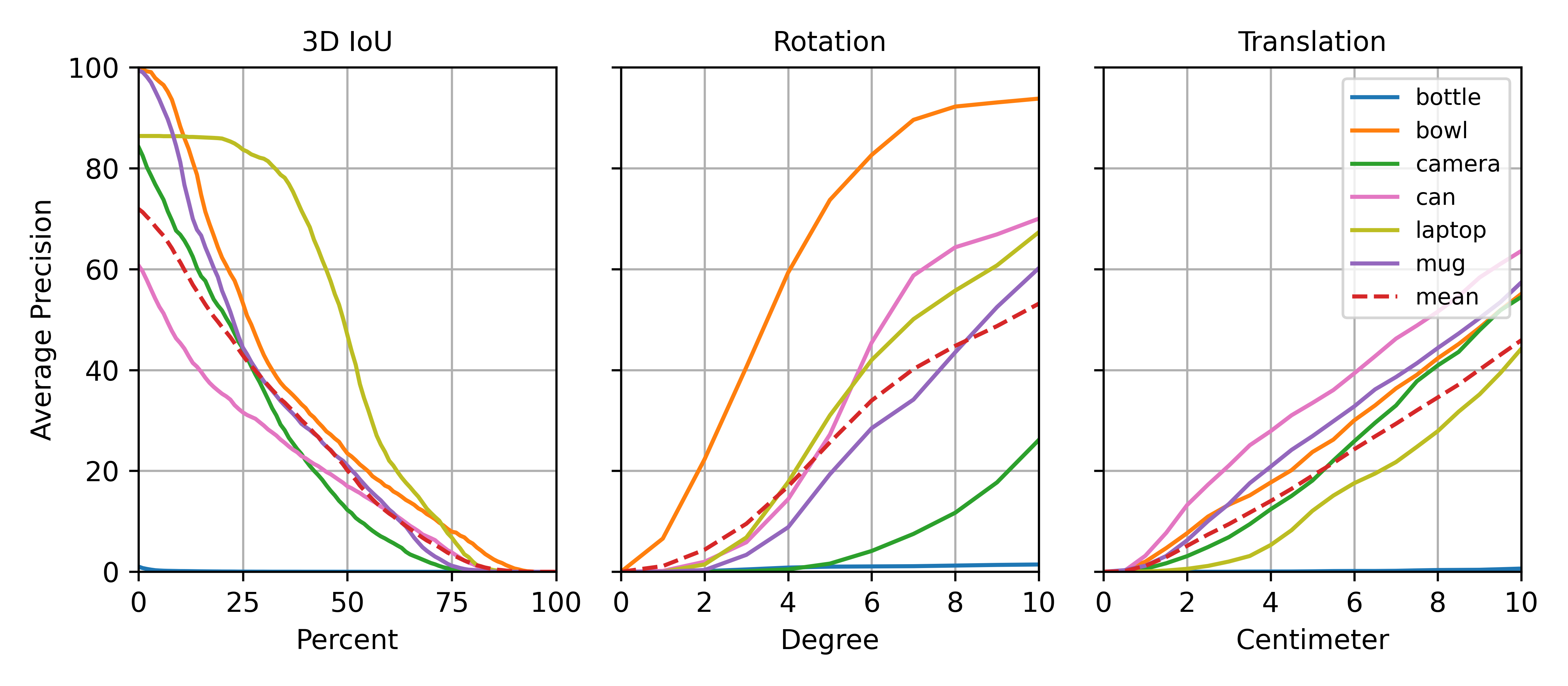} \\
  \includegraphics[height=6cm]{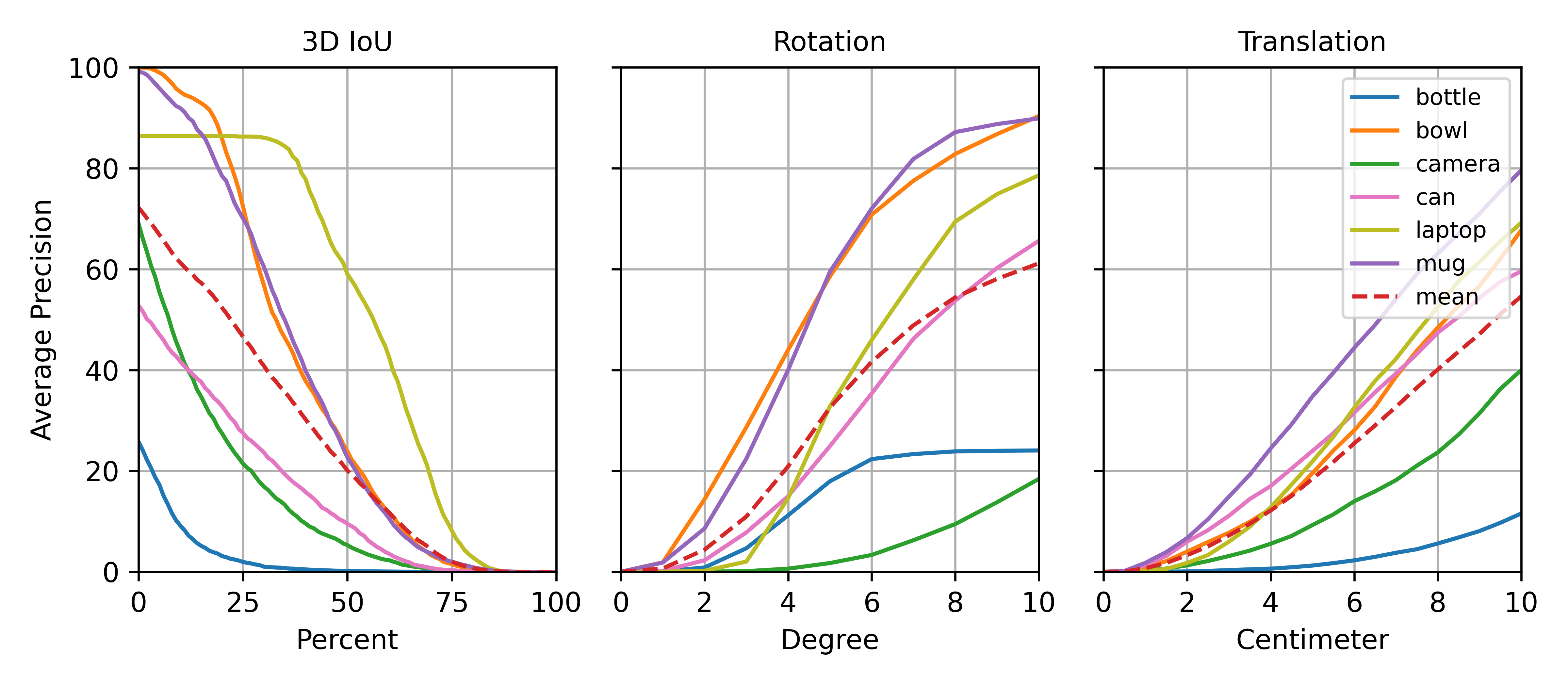}
  \caption{The average precision for various error thresholds for each category on REAL275 for GIVEPose (top) and ours (bottom).}
  \label{fig:ap}
\end{figure}

The category-wise average precision (AP) curves across varying error thresholds are illustrated in Figure~\ref{fig:ap}, comparing GivePose (top) and our proposed framework (bottom). Our method demonstrates consistent performance gains across most object classes, with the most pronounced improvements observed in the \textit{bottle} and \textit{mug} categories. The substantial performance leap in the \textit{bottle} class underscores the model's capacity to simultaneously resolve continuous rotational symmetries and severe scale ambiguities driven by intra-class geometric variations. This heightened rotation accuracy on highly symmetric objects confirms that our ScoreNet successfully preserves multi-modal pose hypotheses rather than suffering from early mode collapse.

\section{Qualitative Results on REAL275}
\label{sec:qual-real}
\begin{figure*}[tb]
  \centering
    \includegraphics[height=15cm]{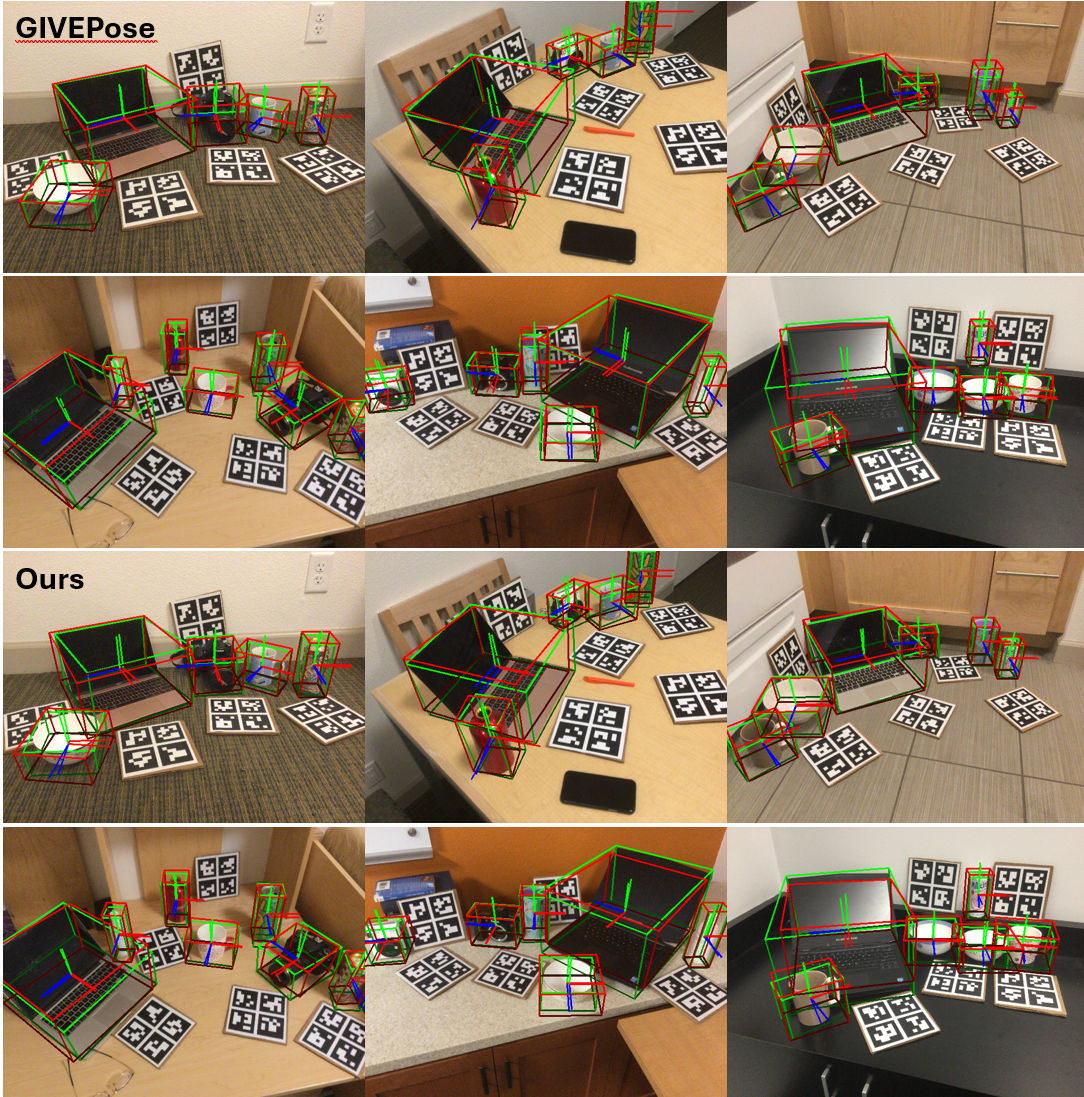}  
  \caption{Further Qualitative results of GIVEPose and our method on the REAL275 dataset with red predicted bounding boxes and green ground truth bounding box.}
  \label{fig:qual-real}
\end{figure*}

Figure \ref{fig:qual-real} illustrates more qualitative results of GIVEPose and our method on the REAL275 dataset. The results show the difference between the ground truth and predicted poses as well as the 3D bounding boxes (green is ground truth and red is predicted). The absence of a predicted bounding box and pose indicates the object was not detected by the 2D object detector.

\end{document}